\documentclass[10pt,twocolumn,letterpaper]{article}

\usepackage{balance}
\usepackage{graphics,color}
\usepackage{cvpr}
\usepackage{times}
\usepackage{epsfig}
\usepackage{graphicx}
\usepackage{amsmath}
\usepackage{amssymb}
\usepackage{multirow}
\usepackage{bm}
\usepackage{booktabs}
\usepackage{stfloats}
\usepackage[square, comma, sort&compress, numbers]{natbib}
\usepackage{subcaption}

\usepackage[pagebackref=false,breaklinks=true,letterpaper=true,colorlinks,bookmarks=false]{hyperref}

\cvprfinalcopy 

\def\cvprPaperID{8239} 

\ifcvprfinal
\begin{document}
\setcitestyle{number}
\title{DoveNet: Deep Image Harmonization via Domain Verification}

\author{$\textnormal{Wenyan Cong}^{1}$, $\textnormal{Jianfu Zhang}^{1}$, $\textnormal{Li Niu}^{1}$\thanks{Corresponding author.}, $\textnormal{Liu Liu}^{1}$, $\textnormal{Zhixin Ling}^{1}$, $\textnormal{Weiyuan Li}^{2}$, $\textnormal{Liqing Zhang}^{1}$\\
$^1$ MoE Key Lab of Artificial Intelligence, Shanghai Jiao Tong University
$^2$ East China Normal University\\
{\tt\small$^1$\{plcwyam17320,c.sis,ustcnewly,Shirlley,1069066484\}@sjtu.edu.cn }\\
{\tt\small$^2$10162100162@stu.ecnu.edu.cn $^1$zhang-lq@cs.sjtu.edu.cn }
}

\maketitle
\thispagestyle{empty}

\begin{abstract}
Image composition is an important operation in image processing, but the inconsistency between foreground and background significantly degrades the quality of composite image. Image harmonization, aiming to make the foreground compatible with the background, is a promising yet challenging task. However, the lack of high-quality publicly available dataset for image harmonization greatly hinders the development of image harmonization techniques. In this work, we contribute an image harmonization dataset iHarmony4 by generating synthesized composite images based on COCO (resp., Adobe5k, Flickr, day2night) dataset, leading to our HCOCO (resp., HAdobe5k, HFlickr, Hday2night) sub-dataset. Moreover, we propose a new deep image harmonization method DoveNet using a novel domain verification discriminator, with the insight that the foreground needs to be translated to the same domain as background. Extensive experiments on our constructed dataset demonstrate the effectiveness of our proposed method. Our dataset and code are available at \href{https://github.com/bcmi/Image\_Harmonization\_Datasets}{https://github.com/bcmi/Image\_Harmonization\_Datasets}.
\end{abstract}


\section{Introduction} \label{sec:intro}

Image composition targets at generating a composite image by extracting the foreground of one image and pasting it on the background of another image. However, since the foreground is usually not compatible with the background, the quality of composite image would be significantly downgraded. To address this issue, image harmonization aims to adjust the foreground to make it compatible with the background in the composite image. Both traditional methods~\cite{lalonde2007using,xue2012understanding,zhu2015learning} and deep learning based method~\cite{tsai2017deep,xiaodong2019improving} have been explored for image harmonization, in which deep learning based method~\cite{tsai2017deep,xiaodong2019improving} could achieve promising results.

As a data-hungry approach, deep learning calls for a large number of training pairs of composite image and harmonized image as input image and its ground-truth output. However, given a composite image, manually creating its harmonized image, \ie, adjusting the foreground to be compatible with background, is in high demand for extensive efforts of skilled expertise. So this strategy of constructing datasets is very time-consuming and expensive, making it infeasible to generate large-scale training data. Alternatively, as proposed in~\cite{tsai2017deep}, we can treat a real image as harmonized image, segment a foreground region, and adjust this foreground region to be inconsistent with the background, yielding a synthesized composite image. Then, pairs of synthesized composite image and real image can be used to supersede pairs of composite image and harmonized image. Because foreground adjustment can be done automatically (\emph{e.g.}, color transfer methods) without time-consuming expertise editing, it becomes feasible to collect large-scale training data. Despite this inspiring strategy proposed in~\cite{tsai2017deep}, Tsai \emph{et al.}~\cite{tsai2017deep} did not make the constructed datasets publicly available. Besides, the proposed dataset has several shortcomings, such as inadequate diversity/realism of synthesized composite images and lack of real composite images.

Considering the unavailability and shortcomings of the dataset built in~\cite{tsai2017deep}, we tend to build our own stronger dataset. Overall, we adopt the strategy in~\cite{tsai2017deep} to generate pairs of synthesized composite image and real image. Similar to~\cite{tsai2017deep}, we generate synthesized composite images based on Microsoft COCO dataset~\cite{lin2014microsoft}, MIT-Adobe5k dataset~\cite{bychkovsky2011learning}, and our self-collected Flickr dataset. For Flickr dataset, we crawl images from Flickr image website by using the category names in ImageNet dataset~\cite{imagenet_cvpr09} as queries in order to increase the diversity of crawled images. Nevertheless, not all crawled images are suitable for the image harmonization task. So we manually filter out the images with pure-color or blurry background, the cluttered images with no obvious foreground objects, and the images which appear apparently unrealistic due to artistic editing.

Besides COCO, Adobe5k, and Flickr suggested in~\cite{tsai2017deep}, we additionally consider datasets which contain multiple images captured in different conditions for the same scene. Such datasets are naturally beneficial for image harmonization task because composite images can be easily generated by replacing the foreground region in one image with the same foreground region in another image. More importantly, two foreground regions are both from real images and thus the composite image is actually a real composite image. However, to the best of our knowledge, there are only a few available datasets~\cite{shih2013data, zhou2016evaluating,Laffont14} within this scope.
Finally, we choose day2night dataset~\cite{Laffont14}, because day2night provides a collection of aligned images captured in a variety of conditions (\emph{e.g.}, weather, season, time of day) for each scene. According to the names of original datasets, we refer to our constructed sub-datasets as HCOCO, HAdobe5k, HFlickr, and Hday2night, with ``H" standing for ``Harmonization". All four sub-datasets comprise a large-scale image harmonization dataset. The details of constructing four sub-datasets and the difference from \cite{tsai2017deep} will be fully described in Section~\ref{sec:data_cons}.

As another contribution, we propose DoveNet, a new deep image harmonization method with a novel domain verification discriminator. Given a composite image, its foreground and background are likely to be captured in different conditions (\emph{e.g.}, weather, season, time of day), and thus have distinctive color and illumination characteristics, which make them look incompatible. Following the terminology in domain adaptation~\cite{patel2015visual,dafv2016} and domain generalization~\cite{wsdg2015,mvdg2015}, we refer to each capture condition as one domain and there could be numerous possible domains. In this case, the foreground and background of a composite image belong to two different domains, while the foreground and background of a real image belong to the same domain. Therefore, the goal of image harmonization, \emph{i.e.}, adjusting the foreground to be consistent with background, can be deemed as translating the domain of foreground to the same one as background without knowing the domain labels of foreground and background. Inspired by adversarial learning~\cite{goodfellow2014generative,pixelGAN}, we propose a domain verification discriminator to pull close the domains of foreground and background in a harmonized image. Specifically, we treat the paired foreground and background representations of a real (\emph{resp.}, composite) image as a positive (\emph{resp.}, negative) pair. On the one hand, we train the discriminator to distinguish positive pairs from negative pairs. On the other hand, the generator is expected to produce a harmonized image, which can fool the discriminator into perceiving its foreground-background pair as positive. To verify the effectiveness of our proposed domain verification discriminator, we conduct comprehensive experiments on our constructed dataset. Our main contributions are summarized as follows:
\begin{itemize}
\item We release the first large-scale image harmonization dataset iHarmony4 consisting of four sub-datasets: HCOCO, HAdobe5K, HFlickr, and Hday2night.
\item We are the first to introduce the concept of domain verification, and propose a new image harmonization method DoveNet equipped with a novel domain verification discriminator.
\end{itemize}

\begin{figure*}	
	\centering
	\begin{subfigure}[t]{3.2in}
		\centering
		\includegraphics[width=8cm]{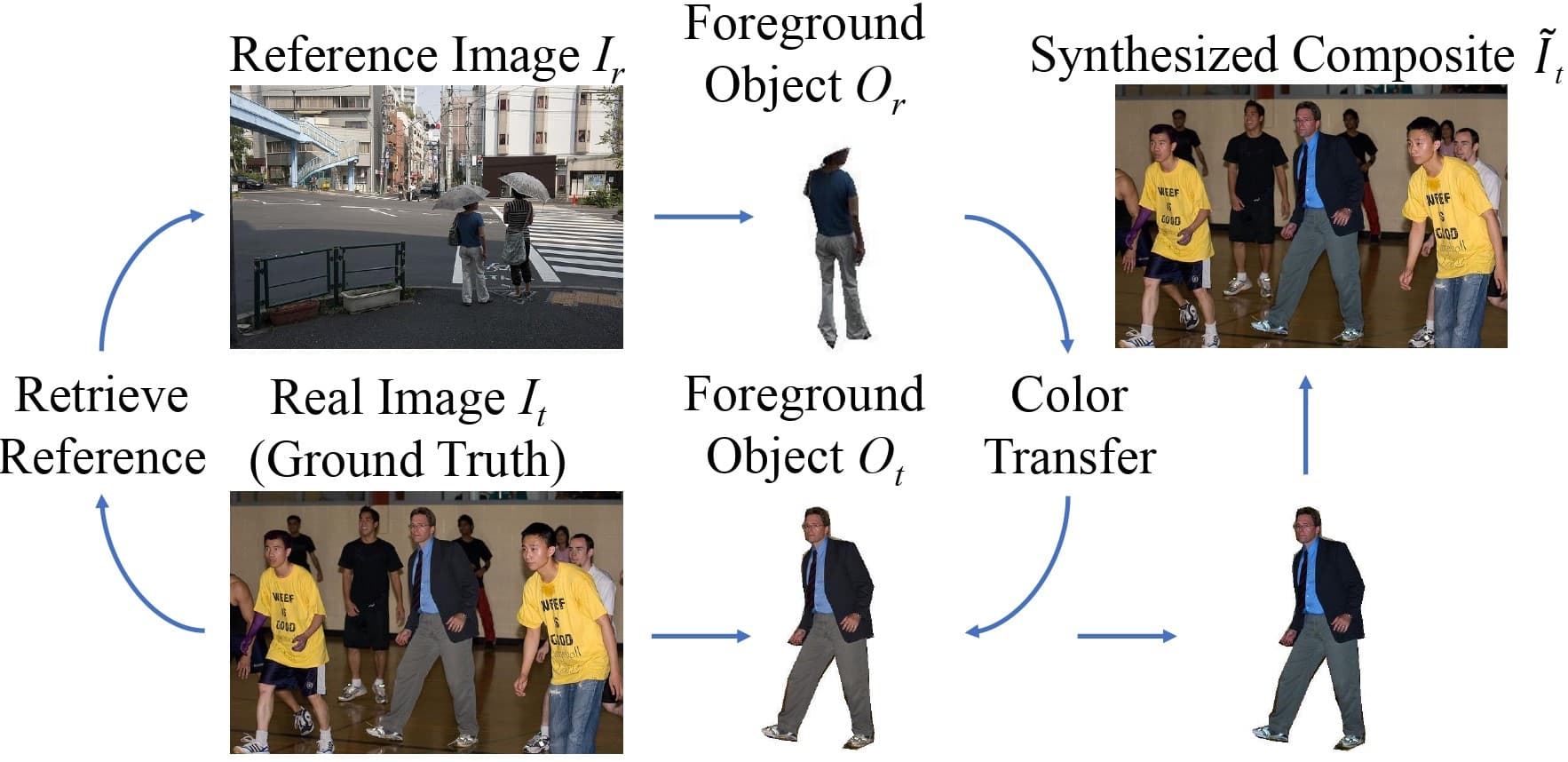}
		\caption{Microsoft COCO $\&$ Flickr}\label{Fig:coco_acq}		
	\end{subfigure}
	\quad
	\begin{subfigure}[t]{3.2in}
		\centering
		\includegraphics[width=8cm]{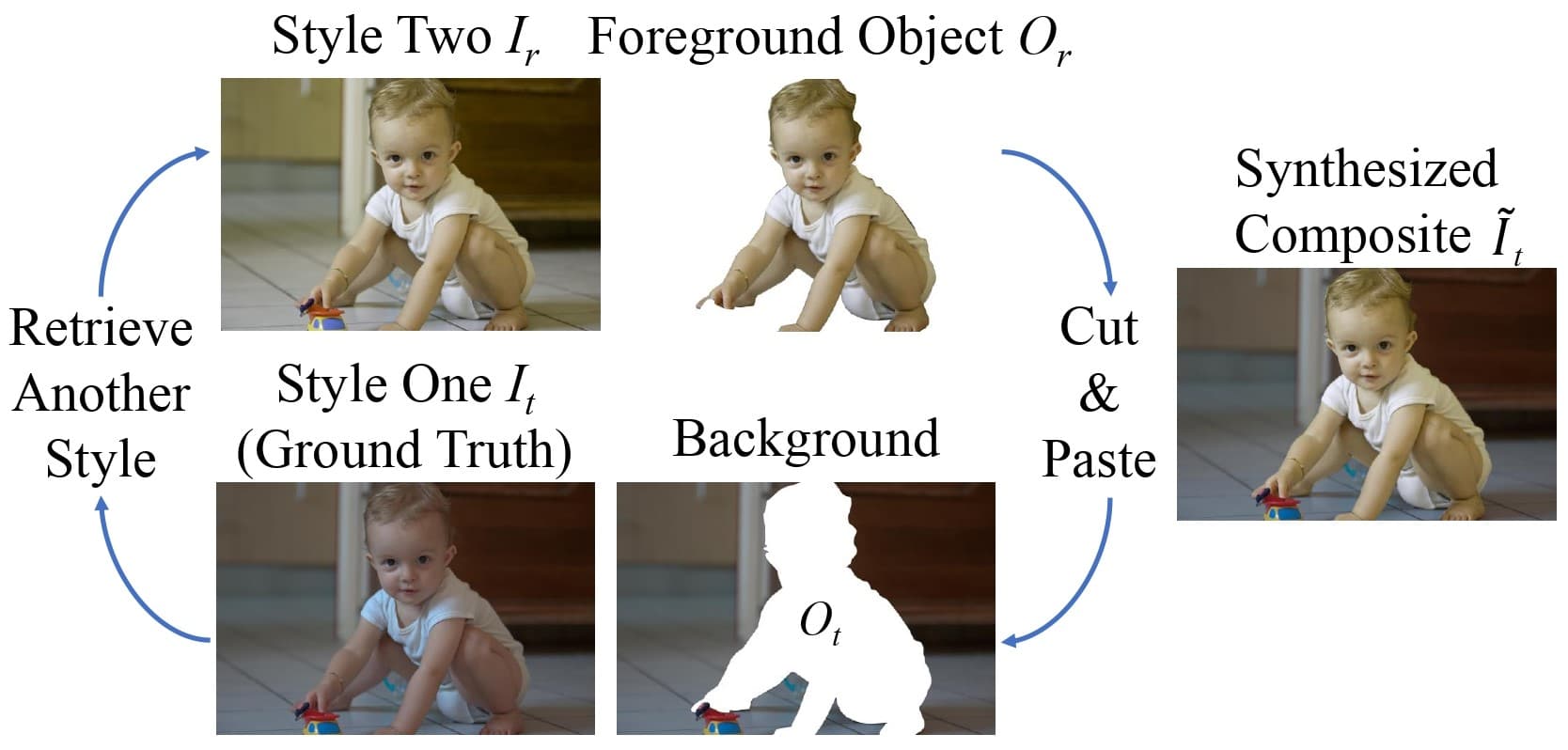}
		\caption{MIT-Adobe Fivek $\&$ day2night}\label{Fig:adobe_acq}
	\end{subfigure}
	\caption{The illustration of our data acquisition process. (a) On Miscrosoft COCO and Flickr datasets, given a target image $I_t$ with foreground object $O_t$, we find a reference image $I_r$ with foreground object $O_r$ from the same category as $O_t$, and then transfer color information from $O_r$ to $O_t$. (b) On MIT-Adobe5k and day2night datasets, given a target image $I_t$ with foreground object $O_t$, we find its another version $I_r$ (edited to present a different style or captured in a different condition) and overlay $O_t$ with the corresponding $O_r$ at the same location in $I_r$.}\label{fig:data_acq}
\end{figure*}

\section{Related Work}
In this section, we review the development of image harmonization. Besides, as image harmonization is a special case of image-to-image translation, we discuss other related applications in this realm.

\textbf{Image Harmonization: }
Traditional image harmonization methods concentrated on better matching low-level appearance statistics, such as matching global and local color distributions \cite{Pitie2005ndimensional,reinhard2001color}, mapping to predefined perceptually harmonious color templates \cite{colorharmonization}, applying gradient-domain compositing \cite{poisson,dragdroppaste,error-tolerant}, and transferring multi-scale various statistics \cite{multi-scale}. To link lower-level image statistics with higher-level properties, visual realism of composite images is further considered in \cite{lalonde2007using,xue2012understanding}. 

Recently, Zhu \emph{et al.} \cite{zhu2015learning} trained a CNN model to perform realism assessment of composite images and applied the model to improve realism. Tsai \emph{et al.} \cite{tsai2017deep} proposed the first end-to-end CNN network to directly produce harmonized images, in which an extra segmentation branch is used to incorporate semantic information. In \cite{xiaodong2019improving}, an attention module was proposed to learn the attended foreground and background features separately. Different from these existing methods, our proposed method aims to translate the foreground domain to the background domain by using a domain verification discriminator.

\textbf{Image-to-Image Translation: }A variety of tasks that map an input image to a corresponding output image are collectively named image-to-image translation, such as image super-resolution~\cite{kim2016super-res,kim2016super-res_recursive,ledig2017super-resolution-gan}, inpainting \cite{inpainting,zhang2019GAIN}, colorization~\cite{zhang2016colorful,larsson2016colorization}, denoising~\cite{image-denoising}, de-blurring \cite{deblurring}, dehazing \cite{ren2016dehaze,dehazeNet}, demo-saicking \cite{demosaicking}, decompression \cite{dong2015compression}, and few-shot image generation \cite{fewshotimage2020}. However, there are still limited deep-learning based research in image harmonization field.

Moreover, several general frameworks of image-to-image translation have also been proposed~\cite{pixelGAN,cGAN,zhang2019multi}. Among them, paired GANs like~\cite{pixelGAN} designed for paired training data can be applied to image harmonization, but they do not consider the uniqueness of image harmonization problem. Our model extends paired GAN with a domain verification discriminator, which goes beyond conventional paired GAN.

\section{Dataset Construction} \label{sec:data_cons}
In this section, we will fully describe the data acquisition process to build our dataset iHarmony4. Based on real images, we first generate composite images and then filter out the unqualified composite images.

\subsection{Composite Image Generation}\label{sec:syn_img_gen}

The process of generating synthesized composite image from a real image has two steps: foreground segmentation and foreground adjustment, as illustrated in Figure~\ref{fig:data_acq}.

\noindent\textbf{Foreground Segmentation: }For COCO dataset, we use the provided segmentation masks for $80$ categories. The other datasets (\emph{i.e.}, Adobe5k, Flickr, and day2night) are not associated with segmentation masks, so we manually segment one or more foreground regions for each image. 

On all four sub-datasets, we ensure that each foreground region occupies a reasonable area of the whole image and also attempt to make the foreground objects cover a wide range of categories.

\noindent\textbf{Foreground Adjustment: }After segmenting a foreground region $O_t$ in one image $I_t$, we need to adjust the appearance of $O_t$. For ease of description, $I_t$ is dubbed as target image. As suggested in \cite{tsai2017deep}, another image $I_r$ containing the foreground region $O_r$ is chosen as reference image. Then, color information is transferred from $O_r$ to $O_t$, leading to a synthesized composite image $\tilde{I}_t$.

For Adobe5k dataset, each real image is retouched by five professional photographers, so one real target image $I_t$ is accompanied by five edited images $\{I_i|_{i=1}^5\}$ in different styles. We could randomly select $I_r$ from $\{I_i|_{i=1}^5\}$ and overlay $O_t$ in $I_t$ with the corresponding region $O_r$ at the same location in $I_r$.

For day2night dataset, each scene is captured in different conditions, resulting in a series of aligned images $\{I_i|_{i=1}^n\}$. Similar to Adobe5k, a target image $I_t$ and a reference image $I_r$ could be randomly selected from $\{I_i|_{i=1}^n\}$, followed by overlaying $O_t$ in $I_t$ with the corresponding region $O_r$ in $I_r$. However, different from Adobe5k, we need to make sure that $O_t$ and $O_r$ are the same object without essential change. For example, moving objects (\emph{e.g.}, person, animal, car) in $I_t$ may move or disappear in $I_r$. Besides, even the static objects (\emph{e.g.} building, mountain) in $I_t$ may be different from those in $I_r$, like building with lights on in $I_t$ while lights off in $I_r$. The above foreground changes come from the objects themselves instead of the capture condition, and thus we exclude those pairs from our dataset.

For COCO and Flickr datasets, since they do not have aligned images, given a target image $I_t$ with foreground $O_t$, we randomly select a reference image $I_r$ with foreground $O_r$ belonging to the same category as $O_t$. For COCO dataset with segmentation annotations for 80 categories, given $I_t$ in COCO, we retrieve $I_r$ from COCO itself. For Flickr dataset without segmentation annotations, we use ADE20K pretrained scene-parsing model \cite{zhou2019semantic} to obtain the dominant category label of $O_t$ and retrieve $I_r$ from ADE20K dataset \cite{zhou2019semantic}. Then, as suggested in \cite{tsai2017deep}, we apply color transfer method to transfer color information from $O_r$ to $O_t$. Nevertheless, the work \cite{tsai2017deep} only utilizes one color transfer method~\cite{lee2016automatic}, which limits the diversity of generated images. Considering that color transfer methods can be categorized into four groups based on parametric/non-parametric and correlated/decorrelated color space, we select one representative method from each group, \emph{i.e.}, parametric method~\cite{reinhard2001color} (\emph{resp.}, \cite{xiao2006color}) in decorrelated (\emph{resp.}, correlated) color space and non-parametric method~\cite{fecker2008histogram} (\emph{resp.}, \cite{pitie2007automated}) in decorrelated (\emph{resp.}, correlated) color space. Given a pair of $O_t$ and $O_r$, we randomly choose one from the above four color transfer methods.

\begin{figure*}[tp!]
\begin{center}
\includegraphics[width=0.9\linewidth]{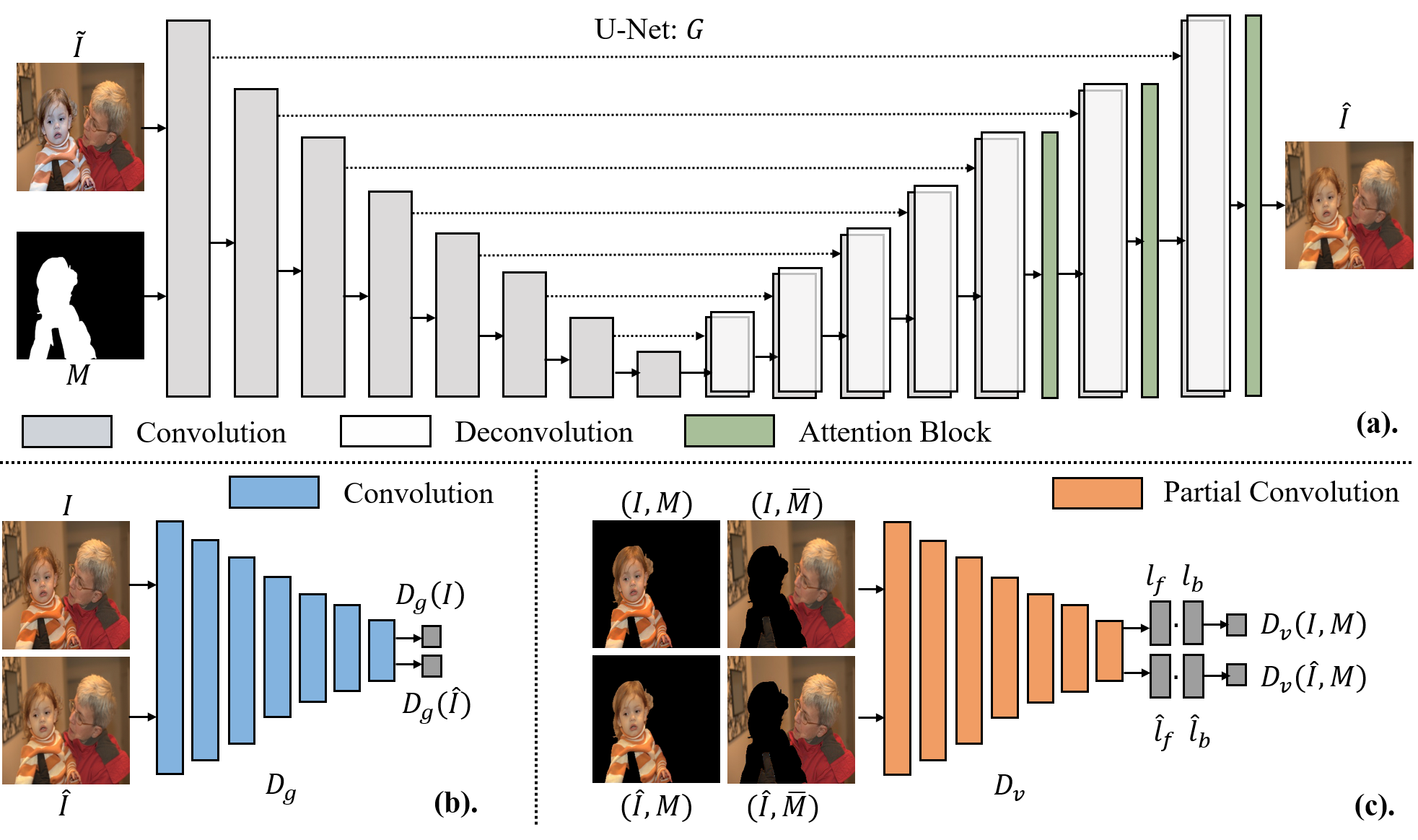}
\end{center}
\caption{Illustration of DoveNet architecture, which consists of (a) attention enhanced U-Net generator, (b) global discriminator, and (c) our proposed domain verification discriminator.}
\label{fig:flowchart}
\end{figure*}

\subsection{Composite Image Filtering} \label{sec:syn_img_filter}
Through foreground segmentation and adjustment, we can obtain a large amount of synthesized composite images. However, some of the synthesized foreground objects look unrealistic, so we use aesthetics prediction model~\cite{kong2016photo} to remove unrealistic composite images. To further remove unrealistic composite images, we train a binary CNN classifier by using the real images as positive samples and the unrealistic composite images identified by~\cite{kong2016photo} as negative samples. When training the classifier, we also feed foreground masks into CNN for better performance.

After two steps of automatic filtering, there are still some remaining unrealistic images. Thus, we ask human annotators to remove the remaining unrealistic images manually. During manual filtering, we also consider another two critical issues: 1) for COCO dataset, some selected foreground regions are not very reasonable such as highly occluded objects, so we remove these images; 2) for COCO and Flickr datasets, the hue of some foreground objects are vastly changed after color transfer, which generally happens to the categories with large intra-class variance. For example, a red car is transformed into a blue car, or a man in red T-shirt is transformed into a man in green T-shirt. This type of color transfer is not very meaningful for image harmonization task, so we also remove these images.

\subsection{Differences between Our Dataset and \cite{tsai2017deep}}
Our dataset iHarmony4 is an augmented and enhanced version of the dataset in \cite{tsai2017deep}:
1) Our dataset contains an additional sub-dataset Hday2night, which is not considered in \cite{tsai2017deep}. Unlike the other three sub-datasets, Hday2night consists of real composite images, which is closer to real-world application; 2) Besides, we also attempt to address some issues not considered in \cite{tsai2017deep}, such as the diversity and quality issues of synthesized composite images; 3) We apply both well-designed automatic filtering and deliberate manual filtering to guarantee the high quality of our dataset.

\section{Our Method}
Given a real image $I$, we have a corresponding composite image $\tilde I$, where the foreground mask $M$ indicates the region to be harmonized and the background mask is $\bar M = 1-M$. Our goal is to train a model that reconstructs $I$ with a harmonized image $\hat I$, which is expected to be as close to $I$ as possible.

We leverage the GAN \cite{goodfellow2014generative} framework to generate plausible and harmonious images. As demonstrated in Figure \ref{fig:flowchart}, in DoveNet, we use an attention enhanced U-Net generator $G$, which takes $(\tilde I, M)$ as inputs and outputs a harmonized image $\hat I$. Besides, we use two different discriminators $D_g$ and $D_v$ to guide $G$ for generating more realistic and harmonious images. The first discriminator $D_g$ is a traditional global discriminator, which discriminates real images and generated images. The second discriminator $D_v$ is our proposed domain verification discriminator, which verifies whether the foreground and background of a given image come from the same domain. 

\subsection{Attention Enhanced Generator}
Our generator $G$ is based on U-Net \cite{ronneberger2015u} with skip links from encoders to decoders. Inspired by \cite{xiaodong2019improving}, we leverage attention blocks to enhance U-Net. Specifically, we first concatenate encoder and decoder features, based on which full attention~\cite{yu2019free} (integration of spatial attention and channel attention) is learnt for encoder feature and decoder feature separately. Then, we concatenate the attended encoder and decoder features. In total, we insert three attention blocks into U-Net as depicted in Figure~\ref{fig:flowchart} and the details of attention block can be found in Supplementary.
We enforce the generated image $\hat I=G(\tilde I, M)$ to be close to ground-truth real image $I$ by $L_{rec} = \|\hat I-I\|_1$.

\subsection{Global Discriminator}
The global discriminator $D_g$ is designed to help $G$ generate plausible images, which takes $I$ as real images and $\hat I$ as fake images. Following \cite{miyato2018spectral}, we apply spectral normalization after each convolutional layer and leverage hinge loss for stabilizing training, which is given by
\begin{equation}\label{eq:gan}
\begin{aligned}
L_{D_g} &= \mathbb{E}[\max(0, 1-D_g(I))] + \mathbb{E}[\max(0, 1+D_g(\hat{I}))] , \\
L_{G_g} &= -\mathbb{E}[D_g(G(\tilde I, M))].
\end{aligned}
\end{equation}
When training $D_g$ by minimizing $L_{D_g}$, $D_g$ is encouraged to produce large (\emph{resp.}, small) scores for real (\emph{resp.}, generated) images. While training $G$ by minimizing $L_{G_g}$, the generated images are expected to fool $D_g$ and obtain large scores.


%


\subsection{Domain Verification Discriminator}
Besides the global discriminator, we also design a domain verification discriminator to verify whether the foreground and background of a given image belong to the same domain. As discussed in Section \ref{sec:intro}, the foreground and background of a real (\emph{resp.}, composite) image are captured in the same condition (\emph{resp.}, different conditions), and thus belong to the same domain (\emph{resp.}, different domains), which is dubbed as a positive (\emph{resp.}, negative) foreground-background pair.

To extract domain representation for foreground and background, we adopt partial convolution \cite{Liu2018}, which is well-tailored for image harmonization task. Partial convolution only aggregates the features from masked regions, which can avoid information leakage from unmasked regions or invalid information corruption like zero padding. 
Our domain representation extractor $F$ is formed by stacking partial convolutional layers, which leverages the advantage of partial convolution to extract domain information for foreground and background separately.

Formally, given a real image $I$, let $I_f=I\circ M$ (\emph{resp.}, $I_b=I\circ \bar M$) be the masked foreground (\emph{resp.}, background) image, in which $\circ$ means element-wise product.
Domain representation extractor $F(I_f,M)$ (\emph{resp.}, $F(I_b, \bar M)$) extracts the foreground representation $l_f$ (\emph{resp.}, $l_b$) based on $I_f$ (\emph{resp.}, $I_b$) and $M$ (\emph{resp.}, $\bar M$). Similarly, given a harmonized image $\hat I$, we apply the same domain representation extractor $F$ to extract its foreground representation $\hat{l}_f$ and background representation $\hat{l}_b$.

After obtaining domain representations, we calculate the domain similarity $D_v(I, M)=l_f\cdot l_b$ (\emph{resp.}, $D_v(\hat I, M)=\hat{l}_f\cdot \hat{l}_b$) as the verification score for the real (\emph{resp.}, generated) images, where $\cdot$ means inner product. In analogy to (\ref{eq:gan}), the loss functions \emph{w.r.t.} the domain verification discriminator can be written as
\begin{equation}
\begin{aligned}
L_{D_v} &= \mathbb{E}[\max(0, 1-D_v(I, M))] \\
&+ \mathbb{E}[\max(0, 1+D_v(\hat{I}, M))], \\
L_{G_v} &= -\mathbb{E}[D_v(G(\tilde I, M), M)].
\end{aligned}
\end{equation}

When training $D_v$ by minimizing $L_{D_v}$, $D_v$ is encouraged to produce large (\emph{resp.}, small) scores for positive (\emph{resp.}, negative) foreground-background pairs. While training $G$ by minimizing $L_{G_v}$, the generated images are expected to fool $D_v$ and obtain large scores. By matching the foreground domain with the background domain, the generated images are expected to have compatible foreground and background. So far, the total loss function for training generator $G$ is
\begin{eqnarray}
L_G=L_{rec}+\lambda (L_{G_g}+ L_{G_v}),
\end{eqnarray}
in which the trade-off parameter $\lambda$ is set as 0.01 in our experiments. Similar to GAN \cite{goodfellow2014generative}, we update generator $G$ and two discriminators $D_g, D_v$ alternatingly. Due to the usage of DOmain VErification (DOVE) discriminator, we name our method as DoveNet.

\setlength{\tabcolsep}{5pt}
\begin{table*}[tb]
\centering
\begin{tabular}{|l|c|c|c|c|c|c|c|c|c|c|}
\hline
\multicolumn{1}{|c|}{Sub-dataset} & \multicolumn{2}{c|}{HCOCO} & \multicolumn{2}{c|}{HAdobe5k} & \multicolumn{2}{c|}{HFlickr} & \multicolumn{2}{c|}{Hday2night} & \multicolumn{2}{c|}{All}\\ \hline
\multicolumn{1}{|c|}{Evaluation metric} & MSE$\downarrow$ & PSNR$\uparrow$ & MSE$\downarrow$ & PSNR$\uparrow$ & MSE$\downarrow$ & PSNR$\uparrow$ & MSE$\downarrow$ & PSNR$\uparrow$ & MSE$\downarrow$ & PSNR$\uparrow$ \\ \hline
\multicolumn{1}{|c|}{input composite} & 69.37 & 33.94 & 345.54 & 28.16 & 264.35 & 28.32 & 109.65 & 34.01 & 172.47 & 31.63 \\ \hline
\multicolumn{1}{|c|}{Lalonde and Efros\cite{lalonde2007using}} & 110.10 & 31.14 & 158.90 & 29.66 & 329.87 & 26.43 & 199.93 & 29.80 & 150.53 & 30.16 \\ \hline
\multicolumn{1}{|c|}{Xue \emph{et al.}\cite{xue2012understanding}} & 77.04 & 33.32 & 274.15 & 28.79 & 249.54 & 28.32 & 190.51 & 31.24 & 155.87 & 31.40 \\ \hline
\multicolumn{1}{|c|}{Zhu \emph{et al.}\cite{zhu2015learning}} & 79.82 & 33.04 & 414.31 & 27.26 & 315.42 & 27.52 & 136.71 & 32.32 & 204.77 & 30.72 \\ \hline
\multicolumn{1}{|c|}{DIH~\cite{tsai2017deep}} & 51.85 & 34.69 & 92.65 & 32.28 & 163.38 & 29.55 & 82.34 & 34.62 & 76.77 & 33.41 \\ \hline
\multicolumn{1}{|c|}{S$^2$AM~\cite{xiaodong2019improving}} & 41.07 & 35.47 & 63.40 & 33.77 & 143.45 & 30.03 & 76.61 & 34.50 & 59.67 & 34.35 \\ \hline
\multicolumn{1}{|c|}{DoveNet} & \bf36.72 & \bf35.83 & \bf52.32 & \bf34.34 & \bf133.14 & \bf30.21 & \bf54.05 & \bf35.18 & \bf52.36 & \bf34.75 \\ \hline
\end{tabular}
\caption{Results of different methods on our four sub-datasets. The best results are denoted in boldface.}
\label{tab:baselines}
\end{table*}

\setlength{\tabcolsep}{4pt}
\begin{table}
\centering
\begin{tabular}{|c|c|c|c|c|}
\hline
Sub-dataset & HCOCO & HAdobe5k & HFlickr & Hday2night\\
\hline
\#Training & 38545 & 19437 & 7449 & 311 \\
\hline
\#Test & 4283 & 2160 & 828 & 133 \\
\hline
\end{tabular}
\caption{The numbers of training and test images on our four sub-datasets.}
\label{tab:statistics}
\vspace{-12pt}
\end{table}

\section{Experiments}
In this section, we analyze the statistics of our constructed iHarmony4 dataset. Then, we evaluate baselines and our proposed DoveNet on our constructed dataset.

\subsection{Dataset Statistics}
\noindent\textbf{HCOCO: }Microsoft COCO dataset ~\cite{lin2014microsoft} contains 118k images for training and 41k for testing. It provides the object segmentation masks for each image with 80 object categories annotated in total. To generate more convincing composites, training set and test set are merged together to guarantee a wider range of available references. Based on COCO dataset, we build our HCOCO sub-dataset with 42828 pairs of synthesized composite image and real image.

\noindent\textbf{HAdobe5k: }MIT-Adobe5k dataset~ \cite{bychkovsky2011learning} covers a wide range of scenes, objects, and lighting conditions. For all the 5000 photos, each of them is retouched by five photographers, producing five different renditions. 
We use 4329 images with one segmented foreground object in each image to build our HAdobe5k sub-dataset, resulting in 21597 pairs of synthesized composite image and real image.

\noindent\textbf{HFlickr: }Flickr website is a public platform for uploading images by amateur photographers.
We construct our HFlickr sub-dataset based on crawled 4833 Flickr images with one or two segmented foreground object in each image. Our HFlickr sub-dataset contains 8277 pairs of synthesized composite image and real image.

\noindent\textbf{Hday2night: }Day2night dataset~\cite{zhou2016evaluating} collected from AMOS dataset~\cite{AMOSdataset} contains images taken at different times of the day with fixed webcams. There are 8571 images of 101 different scenes in total. We select 106 target images from 80 scenes with one segmented foreground object in each image to generate composites. Due to the stringent requirement mentioned in Section~\ref{sec:syn_img_gen}, we only obtain 444 pairs of synthesized composite image and real image, without degrading the dataset quality.

For each sub-dataset (\ie, HCOCO, HAdobe5k, HFlickr, and Hday2night), all pairs are split into training set and test set. We ensure that the same target image does not appear in the training set and test set simultaneously, to avoid that the trained model simply memorize the target image. The numbers of training and test images in four sub-datasets are summarized in Table~\ref{tab:statistics}. The sample images and other statistics are left to Supplementary due to space limitation.

\subsection{Implementation Details}
Following the network architecture in \cite{isola2017image}, we apply eight downsample blocks inside the generator, in which each block contains a convolution with a kernel size of four and stride of two. After the convolution layers, we apply LeakyReLU activation and instance normalization layer. We use eight deconvolution layers to upsample the feature to generate images. For global (\emph{resp.}, verification) discriminator, we use seven convolutional (\emph{resp.}, partial convolutional) layers and LeakyReLU is applied after all the convolutional layers before the last one in both discriminators.
We use Adam optimizer with learning rate 0.002.
Following~\cite{tsai2017deep}, we use Mean-Squared Errors (MSE) and PSNR scores on RGB channels as the evaluation metric. We report the average of MSE and PSNR over the test set.
We resize the input images as $256\times 256$ during both training and testing. MSE and PSNR are also calculated based on $256\times 256$ images.

\subsection{Comparison with Existing Methods}
We compare with both traditional methods~\cite{lalonde2007using,xue2012understanding} and deep learning based methods~\cite{zhu2015learning,tsai2017deep,xiaodong2019improving}. Although Zhu \emph{et al.} \cite{zhu2015learning} is a deep learning based method, it relies on the pretrained aesthetic model and does not require our training set. DIH~\cite{tsai2017deep} originally requires training images with segmentation masks, which are not available in our problem. Therefore, we compare with DIH by removing its semantic segmentation branch, because we focus on pure image harmonization task without using any auxiliary information.
For all baselines, we conduct experiments with their released code if available, and otherwise based on our own implementation.

Following \cite{tsai2017deep}, we merge the training sets of all four sub-datasets as a whole training set to learn the model, which is evaluated on the test set of each sub-dataset and the whole test set. The results of different methods are summarized in Table~\ref{tab:baselines}, from which we can observe that deep learning based methods using our training set~\cite{tsai2017deep,xiaodong2019improving} are generally better than traditional methods~\cite{lalonde2007using,xue2012understanding}, which demonstrates the effectiveness of learning to harmonize images from paired training data. We also observe that S$^2$AM is better than DIH, which shows the benefit of its proposed attention block. Our DoveNet outperforms all the baselines by a large margin and achieves the best results on all four sub-datasets, which indicates the advantage of our domain verification discriminator.

\setlength{\tabcolsep}{5pt}
\begin{table*}[tb]
\centering
\begin{tabular}{|l|c|c|c|c|c|c|c|c|c|c|}
\hline
\multicolumn{1}{|c|}{Sub-dataset} & \multicolumn{2}{c|}{HCOCO} & \multicolumn{2}{c|}{HAdobe5k} & \multicolumn{2}{c|}{HFlickr} & \multicolumn{2}{c|}{Hday2night} & \multicolumn{2}{c|}{All}\\ \hline
\multicolumn{1}{|c|}{Evaluation metric} & MSE$\downarrow$ & PSNR$\uparrow$ & MSE$\downarrow$ & PSNR$\uparrow$ & MSE$\downarrow$ & PSNR$\uparrow$ & MSE$\downarrow$ & PSNR$\uparrow$ & MSE$\downarrow$ & PSNR$\uparrow$ \\ \hline
\multicolumn{1}{|c|}{U-Net} & 46.87 & 34.30 & 77.16 & 32.34 & 160.17 & 29.25 & 57.60 & 34.25 & 68.57 & 33.16 \\ \hline
\multicolumn{1}{|c|}{U-Net+att} & 43.13 & 35.15 & 57.52 & 33.83 & 159.99 & 29.56 & 56.40 & 34.89 & 61.15 & 34.13 \\ \hline
\multicolumn{1}{|c|}{U-Net+att+adv} & 38.44 & 35.54 & 54.56 & 34.08 & 143.03 & 29.99 & 55.68 & 34.72 & 55.15 & 34.48 \\ \hline
\multicolumn{1}{|c|}{U-Net+att+ver} & 39.79 & 35.33 & 53.84 & 34.19 & 136.60 & 30.04 & 55.64 & 34.94 & 55.00 & 34.40 \\ \hline
\multicolumn{1}{|c|}{U-Net+att+adv+ver} & \bf36.72 & \bf35.83 & \bf52.32 & \bf34.34 & \bf133.14 & \bf30.21 & \bf54.05 & \bf35.18 & \bf52.36 & \bf34.75 \\ \hline
\end{tabular}
\caption{Results of our special cases on our four sub-datasets. U-Net is the backbone generator. ``att" stands for our used attention block, ``adv" stands for the adversarial loss of global discriminator. ``ver" stands for the verification loss of our proposed verification discriminator. The best results are denoted in boldface.}
\label{tab:ablate_loss}
\end{table*}

\setlength{\tabcolsep}{8pt}
\begin{table*}[tb]
\centering
\begin{tabular}{|l|c|c|c|c|c|c|c|c|}
\hline
\multicolumn{1}{|c|}{Foreground ratios} & \multicolumn{2}{c|}{$0\%\sim 5\%$} & \multicolumn{2}{c|}{$5\%\sim 15\%$} & \multicolumn{2}{c|}{$15\%\sim 100\%$} & \multicolumn{2}{c|}{$0\%\sim 100\%$}\\ \hline
\multicolumn{1}{|c|}{Evaluation metric} & MSE$\downarrow$ & fMSE$\downarrow$ & MSE$\downarrow$ & fMSE$\downarrow$ & MSE$\downarrow$ & fMSE$\downarrow$ & MSE$\downarrow$ & fMSE$\downarrow$ \\ \hline
\multicolumn{1}{|c|}{Input composite} & 28.51 & 1208.86 & 119.19 & 1323.23 & 577.58 & 1887.05 & 172.47 & 1387.30 \\ \hline
\multicolumn{1}{|c|}{Lalonde and Efros\cite{lalonde2007using}} & 41.52 & 1481.59 & 120.62 & 1309.79 & 444.65 & 1467.98 & 150.53 & 1433.21 \\ \hline
\multicolumn{1}{|c|}{Xue \emph{et al.}\cite{xue2012understanding}} & 31.24 & 1325.96 & 132.12 & 1459.28 & 479.53 & 1555.69 & 155.87 & 1411.40 \\ \hline
\multicolumn{1}{|c|}{Zhu \emph{et al.}\cite{zhu2015learning}} & 33.30 & 1297.65 & 145.14 & 1577.70 & 682.69 & 2251.76 & 204.77 & 1580.17 \\ \hline
\multicolumn{1}{|c|}{DIH~\cite{tsai2017deep}} & 18.92 & 799.17 & 64.23 & 725.86 & 228.86 & 768.89 & 76.77 & 773.18 \\ \hline
\multicolumn{1}{|c|}{S$^2$AM~\cite{xiaodong2019improving}} & 15.09 & 623.11 & 48.33 & 540.54 & 177.62 & 592.83 & 59.67 & 594.67 \\ \hline
\multicolumn{1}{|c|}{DoveNet} & \bf14.03 & \bf591.88 & \bf44.90 & \bf504.42 & \bf152.07 & \bf505.82 & \bf52.36 & \bf549.96 \\ \hline
\end{tabular}
\caption{MSE and foreground MSE (fMSE) of different methods in each foreground ratio range based on the whole test set. The best results are denoted in boldface.}
\label{tab:ablate_ratio}
\end{table*}

\setlength{\tabcolsep}{10pt}
\begin{table}
\centering
\begin{tabular}{|c|c|}
\hline
Method & B-T score$\uparrow$\\\hline
Input composite & 0.624 \\ \hline
Lalonde and Efros \cite{lalonde2007using} & 0.260 \\ \hline
Xue \emph{et al.} \cite{xue2012understanding} & 0.567 \\ \hline
Zhu \emph{et al.} \cite{zhu2015learning} & 0.337 \\ \hline
DIH~\cite{tsai2017deep} & 0.948 \\ \hline
S$^2$AM~\cite{xiaodong2019improving} & 1.229 \\ \hline
DoveNet & \bf1.437 \\ \hline
\end{tabular}
\caption{B-T scores of different methods on $99$ real composite images provided in \cite{tsai2017deep}. }
\vspace{-5pt}
\label{tab:BT_score}
\end{table}

\subsection{Ablation Studies}
In this section, we first investigate the effectiveness of each component in our DoveNet, and then study the impact of foreground ratio on the harmonization performance.

First, the results of ablating each component are reported in Table~\ref{tab:ablate_loss}. By comparing ``U-Net" with DIH in Table~\ref{tab:baselines}, we find that our backbone generator is better than that used in DIH~\cite{tsai2017deep}. We also observe that ``U-Net+att" outperforms ``U-Net", which shows the benefit of using attention block. Another observation is that ``U-Net+att+adv" (\emph{resp.}, ``U-Net+att+ver") performs more favorably than ``U-Net+att", which indicates the advantage of employing global discriminator (\emph{resp.}, our domain verification discriminator).
Finally, our full method, \emph{i.e.}, ``U-Net+att+adv+ver", achieves the best results on all four sub-datasets.

Second, our dataset has a wide range of foreground ratios (the area of foreground over the area of whole image) in which the foreground ratios of most images are in the range of $[1\%,90\%]$ (see Supplementary). Here, we study the impact of different foreground ratios on the harmonization performance. Especially when the foreground ratio is very small, the reconstruction error of background may overwhelm the harmonization error of foreground. Therefore, besides MSE on the whole image, we introduce another evaluation metric: foreground MSE (fMSE), which only calculates the MSE in the foreground region. We divide foreground ratios into three ranges, \emph{i.e.}, $0\%\sim5\%$, $5\%\sim 15\%$, and $15\%\sim100\%$.
We adopt such a partition because more images have relatively small foreground ratios.
Then, we report MSE and fMSE of different methods for each range on the whole test set in Table~\ref{tab:ablate_ratio}. Obviously, MSE increases as the foreground ratio increases.
Based on Table~\ref{tab:ablate_ratio}, DoveNet outperforms all the baselines \emph{w.r.t.} MSE and fMSE in each range of foreground ratios, especially when the foreground ratio is large, which demonstrates the robustness of our method.

\begin{figure*}[t]
\centering
\includegraphics[width=17cm]{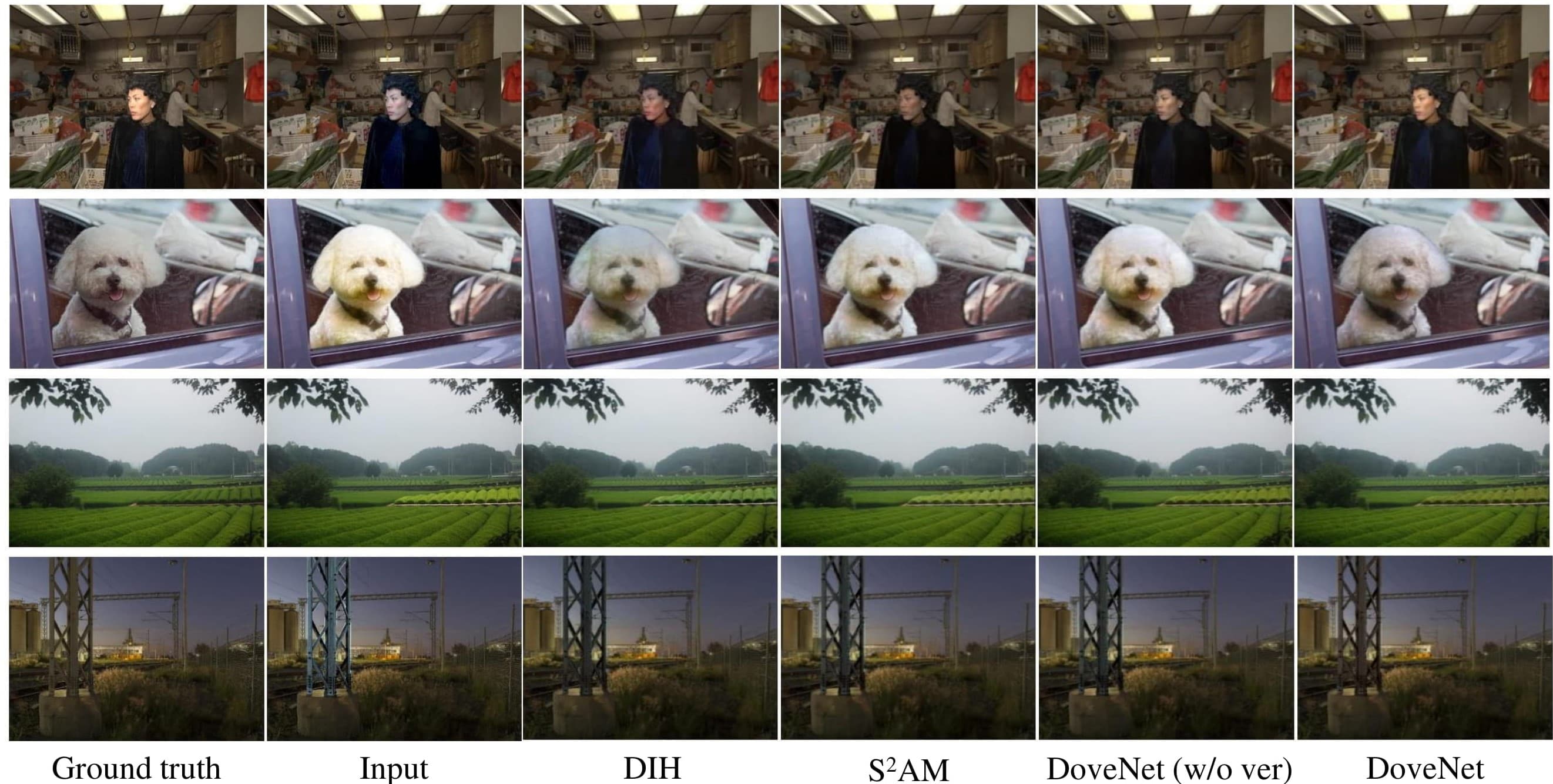}
\caption[]{Example results of different methods on our four sub-datasets. From top to bottom, we show one example from our HAdobe5k, HCOCO, Hday2night, and HFlickr sub-dataset respectively. From left to right, we show the ground-truth real image, input composite image, DIH~\cite{tsai2017deep}, S$^2$AM\cite{xiaodong2019improving}, our special case DoveNet (w/o ver) and our full method DoveNet.}
\label{fig:samples}
\end{figure*}

\subsection{Qualitative Analyses}
In Figure \ref{fig:samples}, we show the ground-truth real image, input composite image, as well as the harmonized images generated by DIH~\cite{tsai2017deep}, S$^2$AM\cite{xiaodong2019improving}, DoveNet (w/o ver), and DoveNet. DoveNet (w/o ver) corresponds to ``U-Net+att+adv" in Table~\ref{tab:ablate_loss}, which removes domain verification discriminator from our method.
We observe that our proposed method could produce the harmonized images which are more harmonious and closer to the ground-truth real images. By comparing DoveNet (w/o ver) and DoveNet, it can be seen that our proposed verification discriminator is able to push the foreground domain close to the background domain, leading to better-harmonized images.

\subsection{User Study on Real Composite Images}
We further compare our proposed DoveNet with baselines on $99$ real composited images used in \cite{tsai2017deep}. Because the provided $99$ real composited images do not have ground-truth images, it is impossible to compare different methods quantitatively using MSE and PSNR. Following the same procedure in \cite{tsai2017deep}, we conduct user study on the $99$ real composited images for subjective evaluation. Specifically, for each real composite image, we can obtain $7$ outputs, including the original composite image and the harmonized images of $6$ methods (see Table~\ref{tab:baselines}). For each real composite image, we can construct pairs of outputs by selecting from $7$ outputs. Then, we invite $50$ human raters to see a pair of outputs at a time and ask him/her to choose the more realistic and harmonious one. A total of $51975$ pairwise results are collected for all $99$ real composite images, in which $25$ results are obtained for each pair of outputs on average. Finally, we use the Bradley-Terry model (B-T model) \cite{bradley1952rank, lai2016comparative} to calculate the global ranking score for each method and report the results in Table \ref{tab:BT_score}.

From Table \ref{tab:BT_score}, we have similar observation as in Table~\ref{tab:baselines}. In particular, deep learning based methods using our training set are generally better than traditional methods, among which DoveNet achieves the highest B-T score. To visualize the comparison, we put the results of different methods on all $99$ real composite images in Supplementary.

\section{Conclusions}
In this work, we have contributed an image harmonization dataset iHarmony4 with four sub-datasets: HCOCO, HAdobe5k, HFlickr, and Hday2night. We have also proposed DoveNet, a novel deep image harmonization method with domain verification discriminator. Extensive experiments on our dataset have demonstrated the effectiveness of our proposed method.

\section*{Acknowledgement}
The work is supported by the National Key R\&D Program of China (2018AAA0100704) and is partially sponsored by National Natural Science Foundation of China (Grant No.61902247) and Shanghai Sailing Program (19YF1424400).

\small
\balance

\pagebreak
\clearpage
\begin{center}
\textbf{\large Supplementary Material for DoveNet: Deep Image Harmonization via Domain Verification}
\end{center}
\setcounter{equation}{0}
\setcounter{section}{0}
\setcounter{figure}{0}
\setcounter{table}{0}
\setcounter{page}{1}
\makeatletter
\renewcommand{\thesection}{S\arabic{section}}
\renewcommand{\theequation}{S\arabic{equation}}
\renewcommand{\thefigure}{S\arabic{figure}}
\renewcommand{\thetable}{S\arabic{table}}
\renewcommand{\bibnumfmt}[1]{[S#1]}
\renewcommand{\citenumfont}[1]{S#1}



In this Supplementary file, we will introduce the attention block used in our proposed DoveNet in Section~\ref{sec:attention}, analyze our constructed iHarmony4 dataset \emph{w.r.t.} foreground ratio, color transfer method, and semantic category in Section~\ref{sec:foreground_ratio}, \ref{sec:color_transfer}, \ref{sec:category}. Besides, We will show samples of manually filtered images and final images in our dataset in Section~\ref{sec:manual_filter}, \ref{sec:dataset_samples}. Finally, we will exhibit the results of different methods on all 99 real composite images in Section~\ref{sec:results_real_composite}.

\section{Details of Attention Block}\label{sec:attention}
\begin{figure*}[tp!]
\begin{center}
\includegraphics[width=0.9\linewidth]{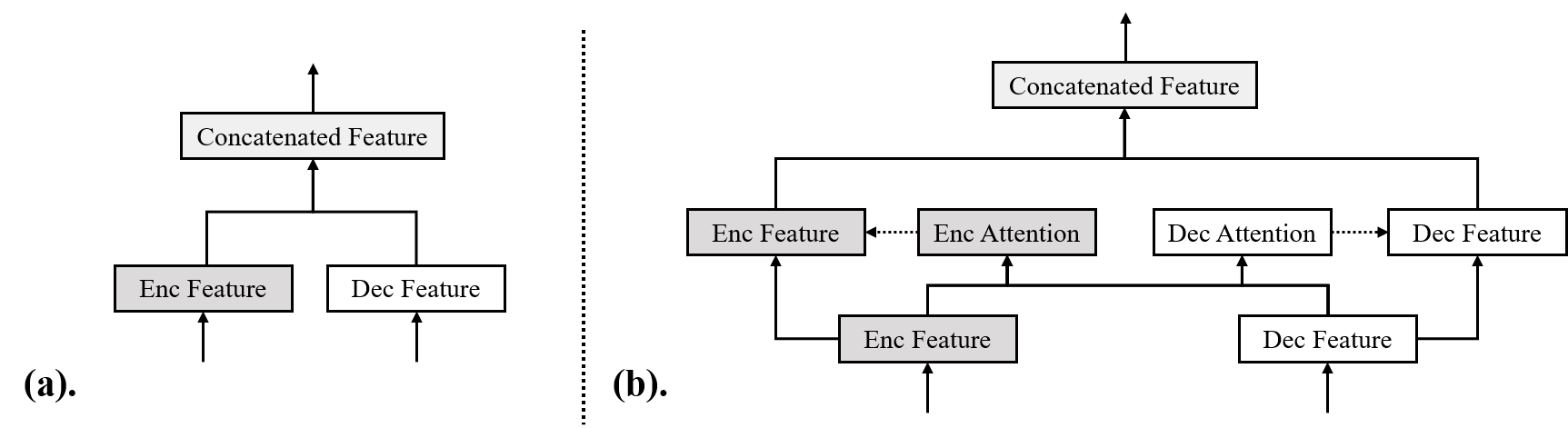}
\end{center}
   \caption{Illustration of our proposed attention module. (a). original U-Net structure without attention module; (b). U-Net with our attention module.}
\label{fig:att}
\end{figure*}

U-Net utilizes skip-connections to leverage information from encoder for decoder. Inspired by \cite{xiaodong2019improvingsupp}, we leverage attention blocks to enhance U-Net.
 The detailed structure is depicted in Fig. \ref{fig:att} (b). We concatenate encoder and decoder features, based on which full attention maps~\cite{yu2019freesupp} (integration of spatial attention and channel attention) are learnt for encoder feature and decoder feature separately. Specifically, to obtain encoder attention map and decoder attention map, we apply $1\times 1$ convolution layer on the concatenation of encoder and decoder features, followed by Sigmoid activation. After that, we multiply encoder (\emph{resp.}, decoder) attention map to the encoder feature (\emph{resp.}, decoder) feature element-wisely. We expect the encoder attention map to pay more attention to the background of encoder feature, because the foreground of encoder feature may not be fully harmonized yet.
 Finally, we concatenate the attended encoder feature and decoder feature as the output of attention block.
 
Note that in \cite{xiaodong2019improvingsupp}, attention maps are learnt for foreground an background separately with explicit mask control. We argue that since the mask is included in the input to the generator, the attention block can utilize the mask information automatically. Compared with \cite{xiaodong2019improvingsupp}, our attention block is simple yet effective.

\section{Analyses of Foreground Ratio}\label{sec:foreground_ratio}

\begin{figure*}[hb]
	\centering
	\begin{subfigure}[t]{3in}
		\centering
		\includegraphics[width=8cm]{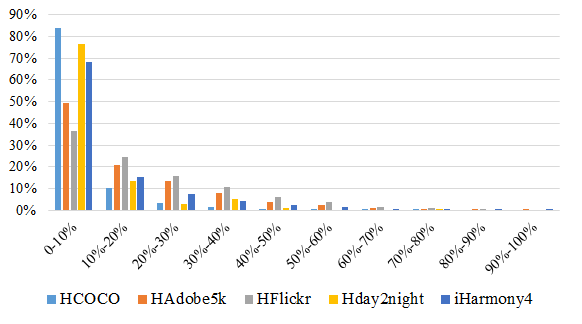}
		\caption{Comparison between four sub-datasets}\label{Fig:R1}		
	\end{subfigure}
	\quad
	\begin{subfigure}[t]{3in}
		\centering
		\includegraphics[width=8cm]{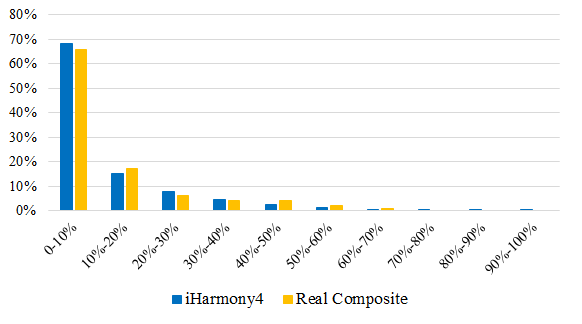}
		\caption{Comparison between iHarmony4 and real composites}\label{Fig:R2}
	\end{subfigure}
	\caption{The distributions of foreground ratios. (a) Comparison between four sub-datasets shows that HCOCO and Hday2night have more images with small foreground regions while HAdobe5k and HFlickr have more images with large foreground regions. (b) Comparison between iHarmony4 and real composites in \cite{xue2012understandingsupp,tsai2017deepsupp} shows that the distribution of foreground ratios of iHarmony4 dataset is close to that of real composite images.}\label{fig:ratio}
\end{figure*}

Our iHarmony4 dataset has a wide range of foreground ratios, \emph{i.e.}, the area of foreground to be harmonized over the area of the whole image. We report the distribution of foreground ratios on our four sub-datasets (\emph{i.e.}, HCOCO, HAdobe5k, HFlickr, Hday2night) and the whole dataset in Figure~\ref{Fig:R1}, from which we can observe that the foreground ratios are mainly distributed in the range $[0\%, 70\%]$ and have a long-tail distribution. We also observe that the foreground ratios on four sub-datasets are quite different, which is caused by different acquisition process of four sub-datasets. Next, we will analyze each sub-dataset separately.

For COCO dataset \cite{lin2014microsoftsupp} with provided segmentation masks, we naturally leverage its segmentation annotation and ensure that each foreground occupies larger than 1\% and smaller than 80\% of the whole image. Since we apply color transfer methods to generate synthesized composites and filter out unqualified ones, synthesized composites with larger foreground regions are more prone to be removed due to more likely low quality. Thus, our HCOCO sub-dataset has relatively small foreground regions.

For the other three datasets (\textit{i.e.}, Adobe5k\cite{bychkovsky2011learningsupp}, Flickr, and day2night\cite{Laffont14supp}), there are no segmentation masks, so we have to manually select and segment one or more foreground regions in an image. The images in Adobe5k are mostly taken by professional photographers with a prominent subject in the image, so it is more likely to select a relatively large foreground. For the crawled Flickr images, we remove those images without obvious foreground and those with blurred background, and thus the remaining images are similar to those in Adobe5k. Therefore, our HAdobe5k and HFlickr sub-datasets have relatively large foreground regions.

For day2night dataset, as discussed in Section 3.1 in the main paper, we need to satisfy more constraints to select a reasonable foreground. Specifically, moving or deformable objects (\emph{e.g.}, person, animal, car) or objects with essential changes are not suitable to be chosen as foreground regions. Hence, we prefer to select static objects that remain consistent across multiple capture conditions, resulting in relatively small foreground regions.

Actually, the composite images in real-world applications also have a wide range of foreground ratios. We show the distribution of 99 real composite images (48 images from Xue \textit{et al.} \cite{xue2012understandingsupp} and 51 images from Tsai \textit{et al.} \cite{tsai2017deepsupp}) and compare with our whole dataset in Figure~\ref{Fig:R2}. From Figure~\ref{Fig:R2}, it can be seen that the distribution of foreground ratios in our whole dataset is close to that of real composite images, which means that the foreground ratios of our constructed dataset are reasonable for real-world image harmonization tasks.

\section{Analyses of Color Transfer Methods}\label{sec:color_transfer}
When constructing HCOCO and HFlickr sub-datasets, we apply color transfer methods to adjust the foreground to make it incompatible with background. 

Existing color transfer methods can be categorized into parametric and non-parametric methods. Parametric methods assume the parametric format of the color mapping function. Assumed parametric format is compact but may be unreliable. Instead, non-parametric methods have no parametric format of the color transfer function, and most of them directly record the mapping of the full range of color/intensity levels using a look-up table, which is usually computed from the 2D joint histogram of image feature correspondences.

From the perspective of color space, existing color transfer methods can be applied to either correlated color space or decorrelated color space. Typically, images are encoded using RGB color space, in which three channels (R, G, B) are highly correlated. This implies that if we want to change the appearance of a pixel in a coherent way, we must modify all three color channels. That complicates any color modification process and may have unpredictable results~\cite{reinhard2001colorsupp}. However, by shifting, scaling, and rotating the axes in RGB color space, we can construct a new color space (\emph{e.g.}, CIELAB, Yuv, HSV). If different channels in this new color space are near-independent or independent, image processing can be done in each channel independently. Nonetheless, decorrelated color space may fail to capture some subtleties including local color information and interrelation~\cite{xiao2006colorsupp}.

Based on the abovementioned parametric or non-parametric methods as well as correlated or decorrelated color space, existing color transfer methods can be grouped into four quadrants, and each quadrant has its own advantage and drawback. To enrich the diversity of synthesized composite images, we choose one representative method in each quadrant.

\textbf{Parametric method in decorrelated color space: }
Based on global color distribution of two images, Reinhard \textit{et al.}~\cite{reinhard2001colorsupp} proposed a linear transformation in decorrelated color space $L\alpha\beta$, by transferring mean and standard deviation between each channel of two images:
\begin{center}
$I_o=\frac{\sigma_r}{\sigma_t}(I_t-\mu_t)+\mu_r$,
\end{center}
where ($\mu_t$, $\sigma_t$) and ($\mu_r$, $\sigma_r$) are the mean and standard deviation of the target and reference images in $L\alpha\beta$ space, $I_t$ and $I_o$ are the color distribution of the target and output images.

\textbf{Parametric method in correlated color space: }
Xiao \textit{et al.} \cite{xiao2006colorsupp} extended \cite{reinhard2001colorsupp} by transferring mean and covariance between images in correlated RGB space. It replaces the rotation to $L\alpha\beta$ with the axes defined by the principal components of each image, leading to a series of matrix transformation:
\begin{center}
$I_o=T_rR_rS_rS_tR_tT_tI_t$,
\end{center}
in which $T,R$ and $S$ denote the matrices of translation, rotation, and scaling derived from the target and reference images accordingly.

\begin{figure*}[hb]
	\centering
	\begin{subfigure}[t]{7in}
		\centering
		\includegraphics[width=16.5cm]{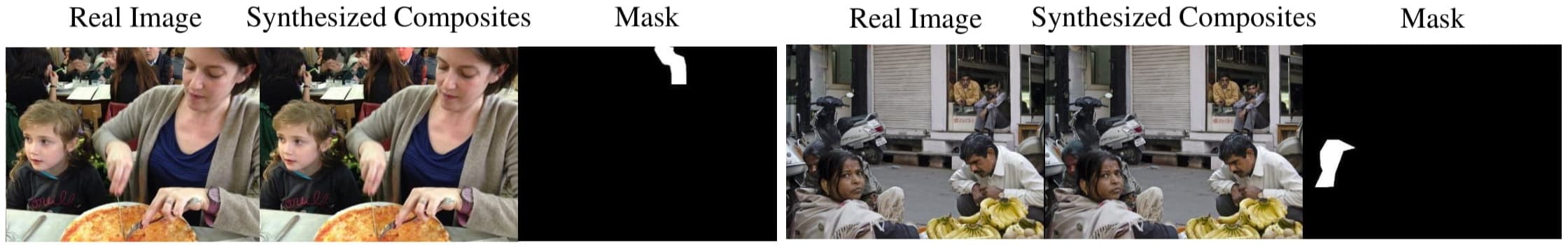}
		\vspace{-5pt}
		\caption{Composite Images with highly occluded foregrounds.}\label{Fig:occluded}
	\end{subfigure}
	\begin{subfigure}[t]{7in}
		\centering
		\includegraphics[width=16.5cm]{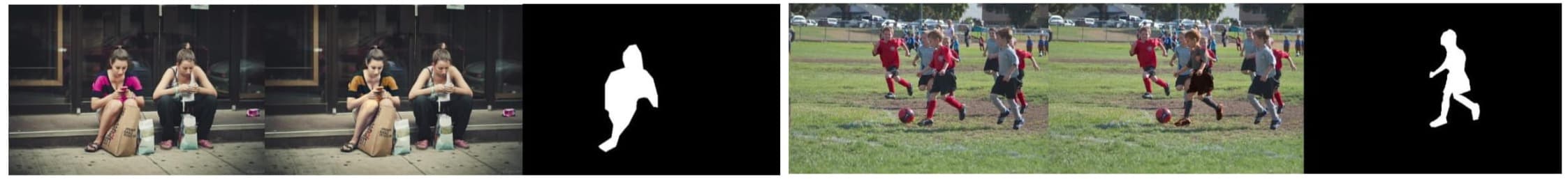}
		\vspace{-5pt}
		\caption{Composite Images with dramatically changed hues.}\label{Fig:hue}
	\end{subfigure}
	\begin{subfigure}[t]{7in}
		\centering
		\includegraphics[width=16.5cm]{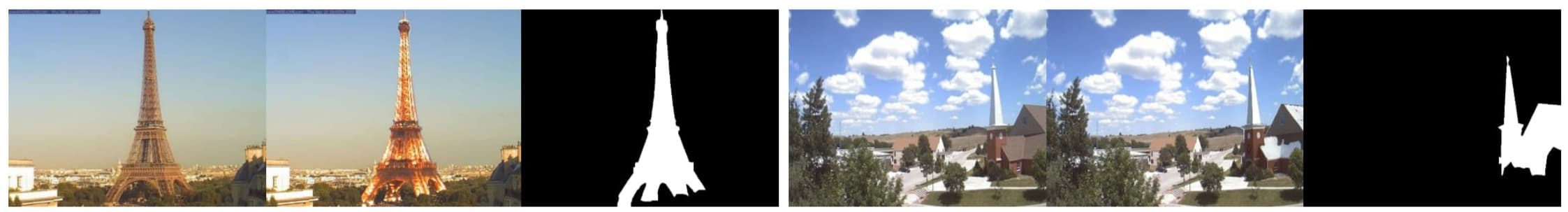}
		\vspace{-5pt}
		\caption{Composite Images with essential object changes.}\label{Fig:object}
	\end{subfigure}

	\caption{Sample composite images that are discarded during manual filtering. From top to bottom, we show undesirable examples with highly occluded foregrounds, dramatically changed hues, and essential changes of the objects themselves that are not caused by capture condition.}
	\label{fig:badsample}
\end{figure*}

\newcommand{\tabincell}[2]{\begin{tabular}{@{}#1@{}}#2\end{tabular}}

\setlength{\tabcolsep}{8pt}
\begin{table*}[tb]
\centering
\begin{tabular}{|l|c|c|c|c|c|c|c|c|}
\hline
\multicolumn{1}{|c|}{Color Transfer Methods} & \multicolumn{2}{c|}{\cite{reinhard2001colorsupp}} & \multicolumn{2}{c|}{\tabincell{c}{\cite{xiao2006colorsupp}}} & \multicolumn{2}{c|}{\tabincell{c}{\cite{pitie2007automatedsupp}}}  & \multicolumn{2}{c|}{\tabincell{c}{\cite{fecker2008histogramsupp}}}\\ \hline
\multicolumn{1}{|c|}{Evaluation metric}  & MSE$\downarrow$  & PSNR$\uparrow$   & MSE$\downarrow$  & PSNR$\uparrow$ & MSE$\downarrow$  & PSNR$\uparrow$ & MSE$\downarrow$  & PSNR$\uparrow$   \\ \hline
\multicolumn{1}{|c|}{Input composite}  & 75.81  & 33.39  & 66.84  & 33.90  & 73.83  & 33.73  & 57.55  & 34.92\\ \hline
\multicolumn{1}{|c|}{Lalonde and Efros~\cite{lalonde2007usingsupp}}  & 127.82 & 30.83 & 107.44 & 31.26 & 110.88 & 31.04 & 92.02 & 31.51 \\ \hline
\multicolumn{1}{|c|}{Xue \emph{et al.}~\cite{xue2012understandingsupp}}  & 90.31 & 32.74 & 69.97 & 33.78 & 74.74 & 33.26 & 72.78 & 33.62 \\ \hline
\multicolumn{1}{|c|}{Zhu \emph{et al.}~\cite{zhu2015learningsupp}}  & 84.31 & 32.62 & 77.93 & 33.06 & 85.36 & 32.78 & 67.81 & 33.89 \\ \hline
\multicolumn{1}{|c|}{DIH~\cite{tsai2017deepsupp}}  & 57.77 & 34.32 & 51.02 & 34.70 & 53.90 & 34.63 & 42.91 & 35.21  \\ \hline
\multicolumn{1}{|c|}{S$^2$AM~\cite{xiaodong2019improvingsupp}}  & 41.89 & 35.29 & 37.33 & 35.81 & 47.00 & 34.99 & 33.93 & 36.13 \\ \hline
\multicolumn{1}{|c|}{Ours}  & \bf38.21 & \bf35.62 & \bf34.42 & \bf36.17 & \bf40.92 & \bf35.38 & \bf30.51 & \bf36.47 \\ \hline
\end{tabular}
\caption{MSE and PSNR on four sub test sets of HCOCO corresponding to different color transfer methods. The best results are denoted in boldface.}
\label{tab:colormapping}
\end{table*}

\textbf{Non-parametric method in decorrelated color space: }
Fecker \textit{et al.} \cite{fecker2008histogramsupp} proposed to use cumulative histogram mapping in decorrelated color space $YC_{b} C_{r}$. They used nearest neighbor mapping scheme to set the corresponding color level of the source image to each level of the target. In this way, the shape of the target histogram can be matched to the reference histogram, and thus the transferred image has the same color as the reference.

\textbf{Non-parametric method in correlated color space: }By treating 3D color distribution as a whole, Pitié \textit{et al.}~\cite{Pitie2005ndimensionalsupp} proposed iterative color distribution transfer by matching 3D distribution through an iterative match of 1D projections. Iterative color distribution transfer can increase the graininess of the original image, especially if the color dynamics of two images are very different. So Pitié \textit{et al.}~\cite{pitie2007automatedsupp} proposed a second stage to reduce the grain artifact through an efficient post-processing algorithm.

In summary, we adopt four color transfer methods: global color transfer in $L\alpha\beta$ space \cite{reinhard2001colorsupp}, global color transfer in RGB space \cite{xiao2006colorsupp}, cumulative histogram matching \cite{fecker2008histogramsupp}, and iterative color distribution transfer \cite{pitie2007automatedsupp}. When generating synthetic composite images for HCOCO and HFlickr, we randomly choose one color transfer method from the above.
By taking HCOCO dataset as an example, after automatic and manual filtering, the number of remaining composite images obtained using method~\cite{reinhard2001colorsupp}~\cite{xiao2006colorsupp}~\cite{pitie2007automatedsupp}~\cite{fecker2008histogramsupp} are 9581, 8119, 17009, and 8119 respectively, which indicates that iterative color distribution transfer \cite{pitie2007automatedsupp} is better at producing realistic and reasonable composites. 

We split the test set of HCOCO into four subsets according to four color transfer methods, and report the results on four subsets in Table~\ref{tab:colormapping}. We can see that the statistics (MSE, PSNR) of input composites obtained by different color transfer methods are considerably different, which shows the necessity of applying multiple color transfer methods to enrich the diversity of synthesized composite images. Moreover, our proposed method achieves the best results on four subsets, which demonstrates the robustness of our proposed method.

\section{Analyses of Semantic Category} \label{sec:category}

COCO dataset is associated with semantic segmentation masks, so we can easily obtain the category labels of foreground regions in our HCOCO sub-dataset. To explore the difference between different categories, we report fMSE (foreground MSE) of input composites and our harmonized results on different categories in Table~\ref{table:category}, in which Table~\ref{tab:hard} (\emph{resp.},~\ref{tab:easy}) shows the hard (\emph{resp.}, easy) categories. We define easy or hard categories based on the relative improvement of our method compared with input composite.

From Table~\ref{table:category}, we find that for categories with small intra-category variance (\emph{e.g.}, mouse, keyboard), fMSE could be improved significantly, while for categories with large intra-category variance (\emph{e.g.}, person), the improvement is relatively small.

\begin{figure*}[hb]
	\centering
	\begin{subfigure}[t]{7in}
		\centering
		\includegraphics[width=16.5cm]{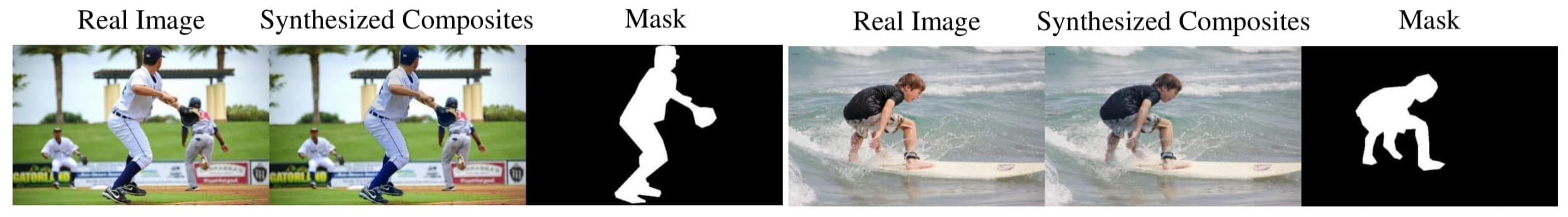}
		\vspace{-5pt}
		\caption{Example images of HCOCO sub-dataset}\label{Fig:hcoco}
	\end{subfigure}
	\begin{subfigure}[t]{7in}
		\centering
		\includegraphics[width=16.5cm]{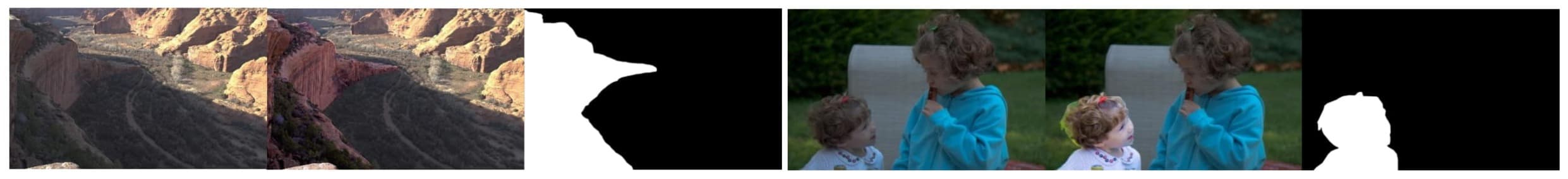}
		\vspace{-5pt}
		\caption{Example images of HAdobe5k sub-dataset}\label{Fig:adobe}
	\end{subfigure}
	\begin{subfigure}[t]{7in}
		\centering
		\includegraphics[width=16.5cm]{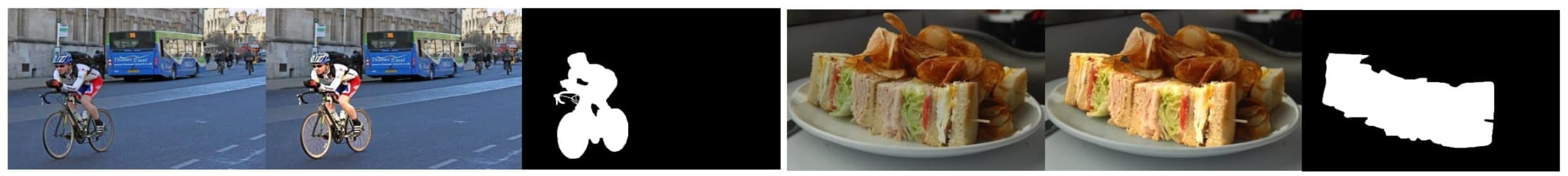}
		\vspace{-5pt}
		\caption{Example images of HFlickr sub-dataset}\label{Fig:flickr}
	\end{subfigure}
	\begin{subfigure}[t]{7in}
		\centering
		\includegraphics[width=16.5cm]{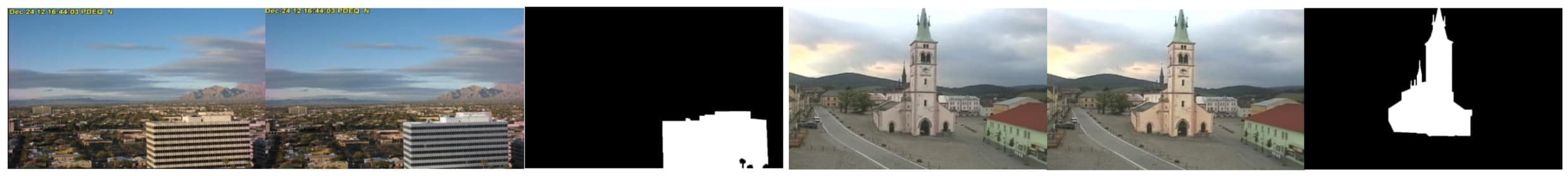}
		\vspace{-5pt}
		\caption{Example images of Hday2night sub-dataset}\label{Fig:day2night}
	\end{subfigure}
	\caption{Example images of our contributed dataset iHarmony4. From top to bottom, we show examples from our HCOCO, HAdobe5k, HFlickr, and Hday2night sub-datasets. From left to right, we show the real image, the synthesized composite image, and the foreground mask for each example.}
	\label{fig:dataset_example}
\end{figure*}

\section{Examples of Manual Filtering} \label{sec:manual_filter}

After generating composite images, two steps of automatic filtering and an additional manually filtering are applied to HCOCO, HFlickr, and Hday2night sub-datasets to eliminate low-quality synthesized composites. In the step of manual filtering, we pay close attention to the cases that are harmful to image harmonization task.

For HCOCO sub-dataset, foreground regions are obtained based on the semantic segmentation annotation provided in COCO dataset. However, some foreground regions are highly occluded and not very meaningful for image harmonization task. For example, in Figure~\ref{Fig:occluded}, an occluded person with only a hand or a shoulder is annotated as ``person'' in COCO dataset, but it is not very meaningful to harmonize a highly occluded person.

Besides, for HCOCO and HFlickr sub-datasets, color transfer is applied to different foreground objects of the same category between reference image and target image, so the hue of foreground object may be dramatically changed, especially for the categories with large intra-category variance like ``person''. For example, in Figure~\ref{Fig:hue}, the shirt color is changed from red in the real image to yellow in the composite image. It does not make sense to harmonize a yellow shirt into a red shirt, so we remove such images from our dataset.

Furthermore, for Hday2night sub-dataset, when overlaying the foreground from reference image on the target image, some essential changes which are not caused by capture condition may happen to the foreground object. For example, in Figure~\ref{Fig:object}, the lights of Eiffel Tower are switched on (see left subfigure) and the snow appears on the roof (see right subfigure) in the composite image. We argue that these changes do not belong to the scope of image harmonization task, and thus filter out these images from our dataset.

By filtering out the unqualified images in the above cases, we ensure the high quality of our iHarmony4 dataset to the utmost.

\setlength{\tabcolsep}{4pt}
\begin{table}	
	\centering
	\hspace{-18pt}
	\begin{subtable}[t]{3in}
		\centering
		\begin{tabular}{|c|c|c|c|c|c|}
			\hline
			  & \tabincell{c}{baseball\\ glove}  & \tabincell{c}{snow-\\board} & kite & person  & \tabincell{c}{surf-\\board}\\
			\hline
			Input  & 1402.62 & 570.07 & 2391.73 & 428.59 & 1409.41\\
			\hline
			DoveNet  & 1301.21 & 523.18 & 1857.24 & 321.48 & 972.09\\
			\hline
		\end{tabular}
		\caption{Categories with slightest fMSE improvement.}
		\label{tab:hard}
	\end{subtable}
	
	\begin{subtable}[t]{3in}
		\hspace{-15pt}
		\begin{tabular}{|c|c|c|c|c|c|}
			\hline
			 & mouse & keyboard & oven & pizza & zebra  \\
			\hline
			Input & 1530.63 & 1624.38 & 1257.21 & 1316.40 & 959.08  \\
			\hline
			DoveNet & 423.84 & 521.24 & 481.37 & 532.38 & 388.66  \\
			\hline
		\end{tabular}
		\caption{Categories with largest fMSE improvement.}
		\label{tab:easy}
	\end{subtable}
	\caption{fMSE improvement of different categories of HCOCO sub-dataset.}\label{table:category}
\end{table}

\section{Examples of Our iHarmony4 Dataset}\label{sec:dataset_samples}

In Figure~\ref{fig:dataset_example}, we show some examples of our four sub-datasets with each row corresponding to one sub-dataset. For each example, we show the original real image, synthesized composite image, and foreground mask. 

\section{Results on Real Composite Images} \label{sec:results_real_composite}

\begin{figure*}[hb]
\begin{center}
\includegraphics[width=\linewidth]{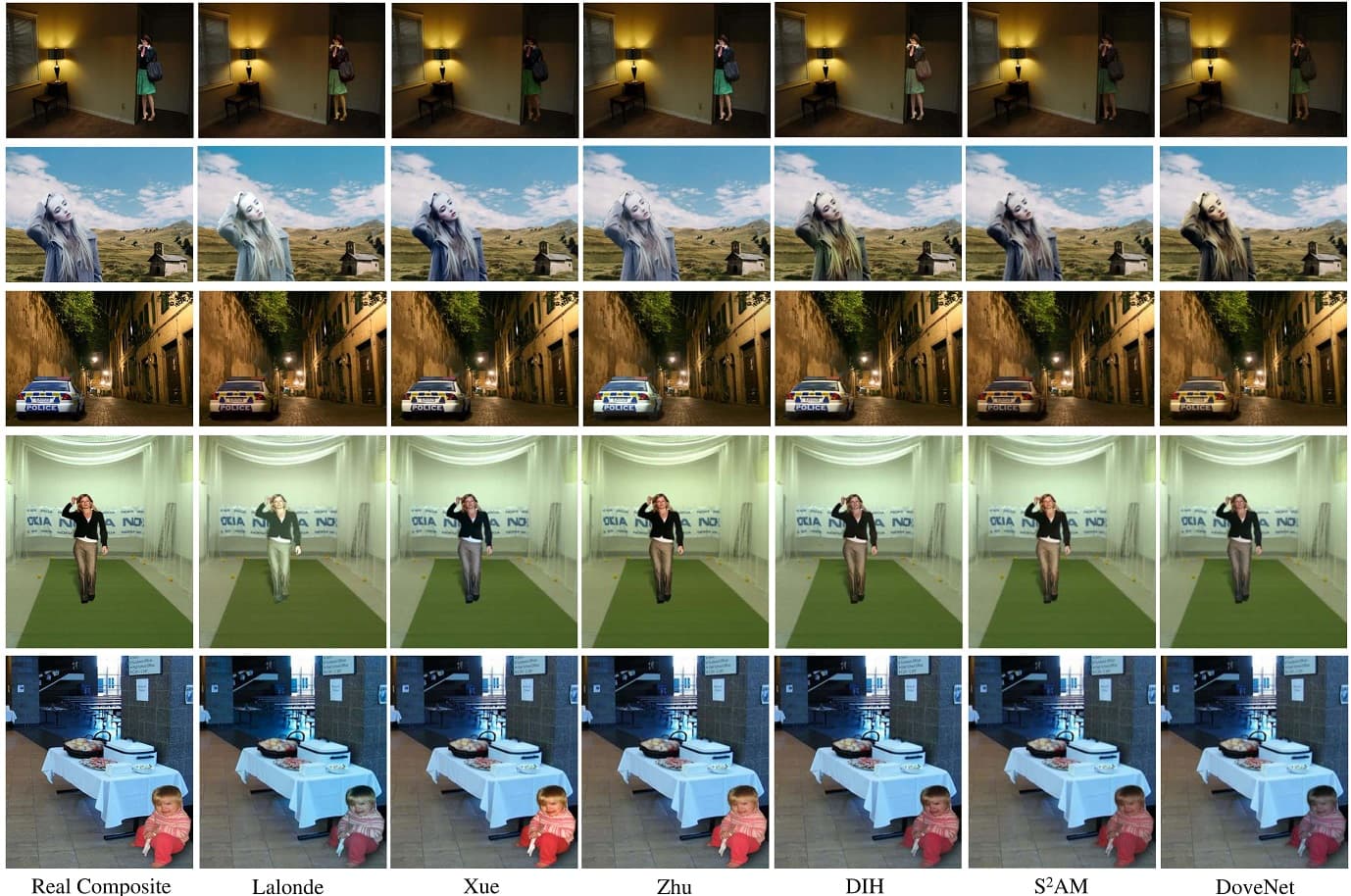}
\end{center}
   \caption{Results on real composite images, including the input composite, five state-of-the-art methods, and our proposed DoveNet.}
\label{fig:real0}
\end{figure*}

\begin{figure*}[tp!]
\begin{center}
\includegraphics[width=\linewidth]{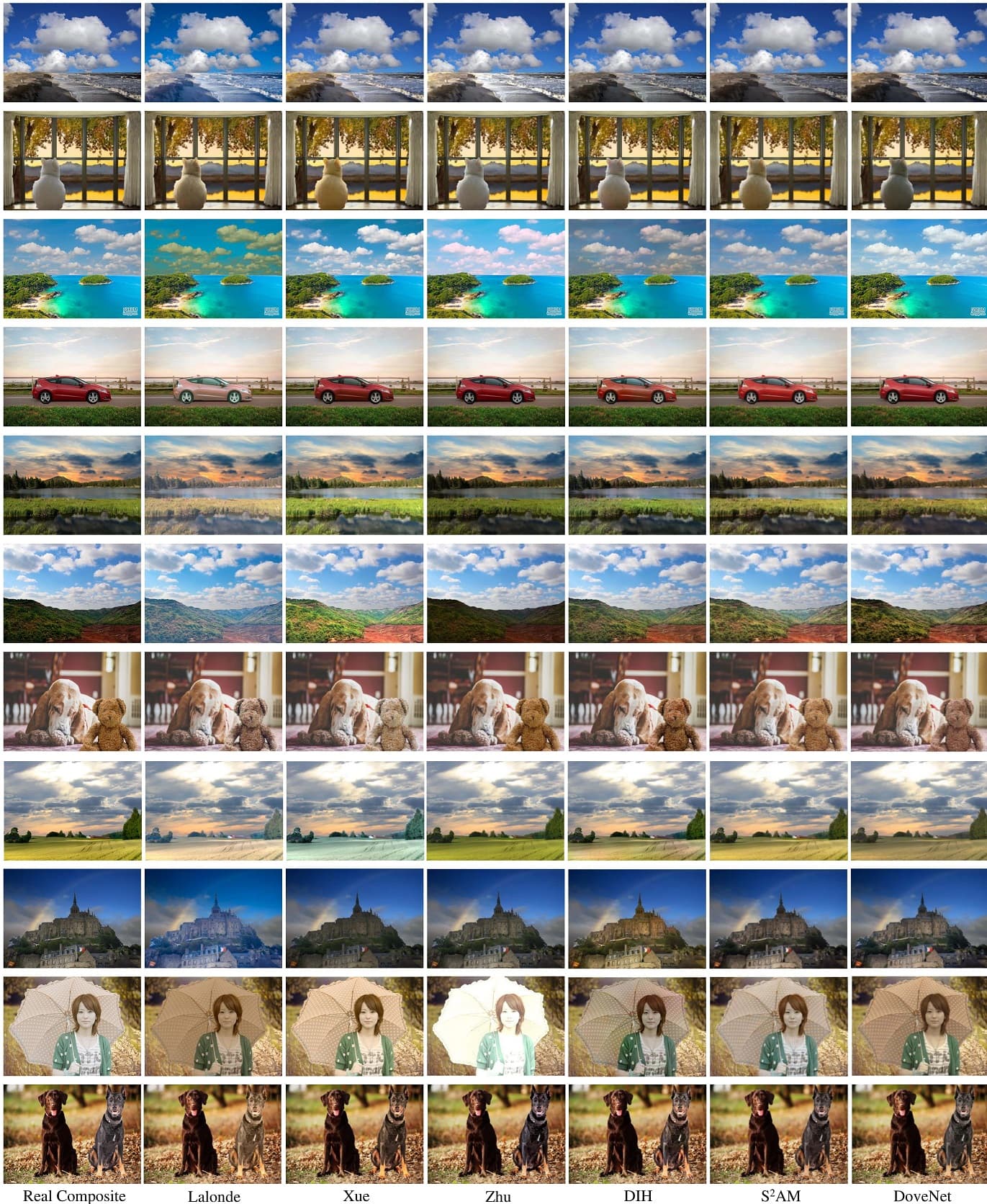}
\end{center}
   \caption{Results on real composite images, including the input composite, five state-of-the-art methods, and our proposed DoveNet.}
\label{fig:real1}
\end{figure*}

\begin{figure*}[tp!]
\begin{center}
\includegraphics[width=\linewidth]{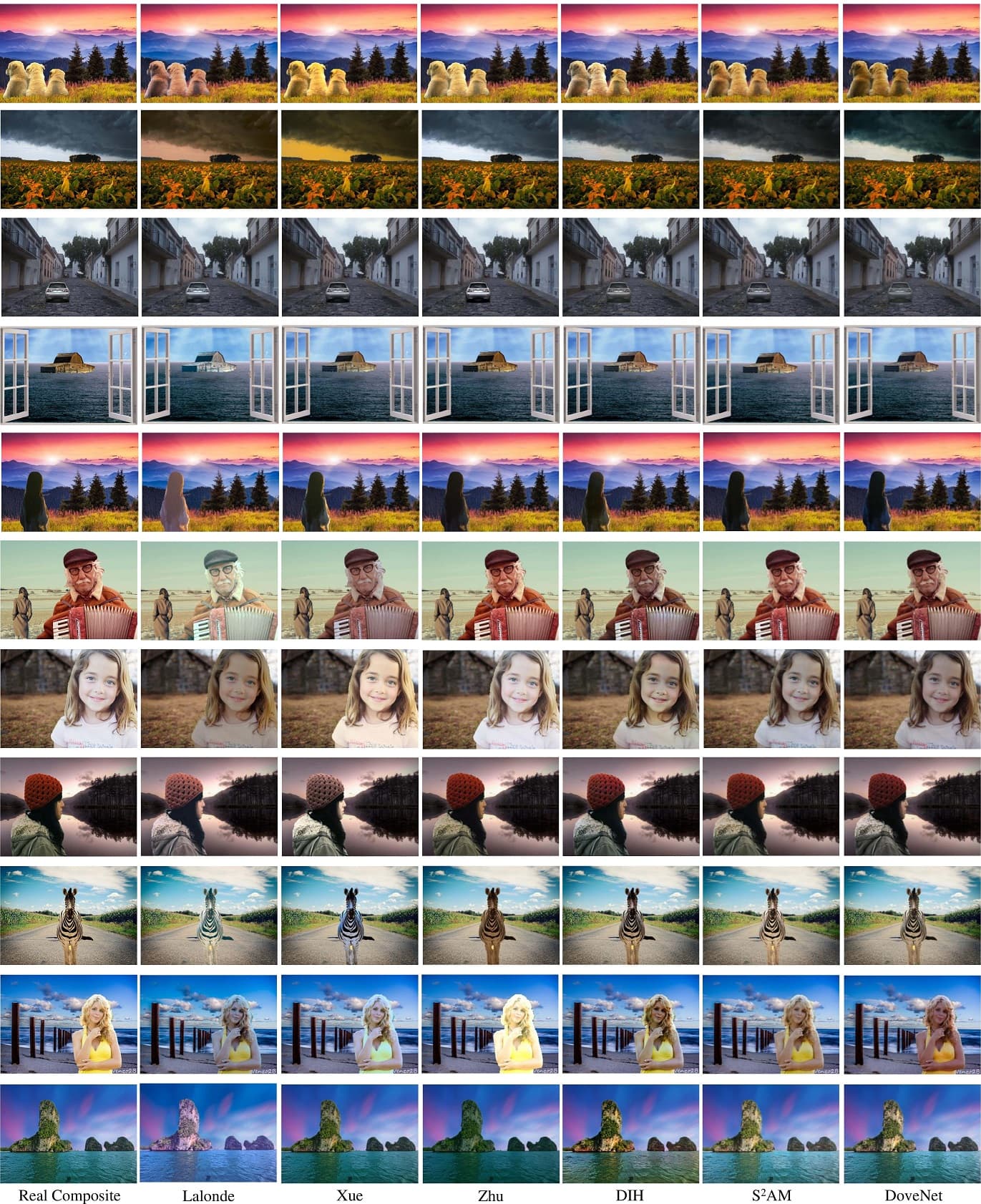}
\end{center}
   \caption{Results on real composite images, including the input composite, five state-of-the-art methods, and our proposed DoveNet. }
\label{fig:real2}
\end{figure*}

\begin{figure*}[tp!]
\begin{center}
\includegraphics[width=\linewidth]{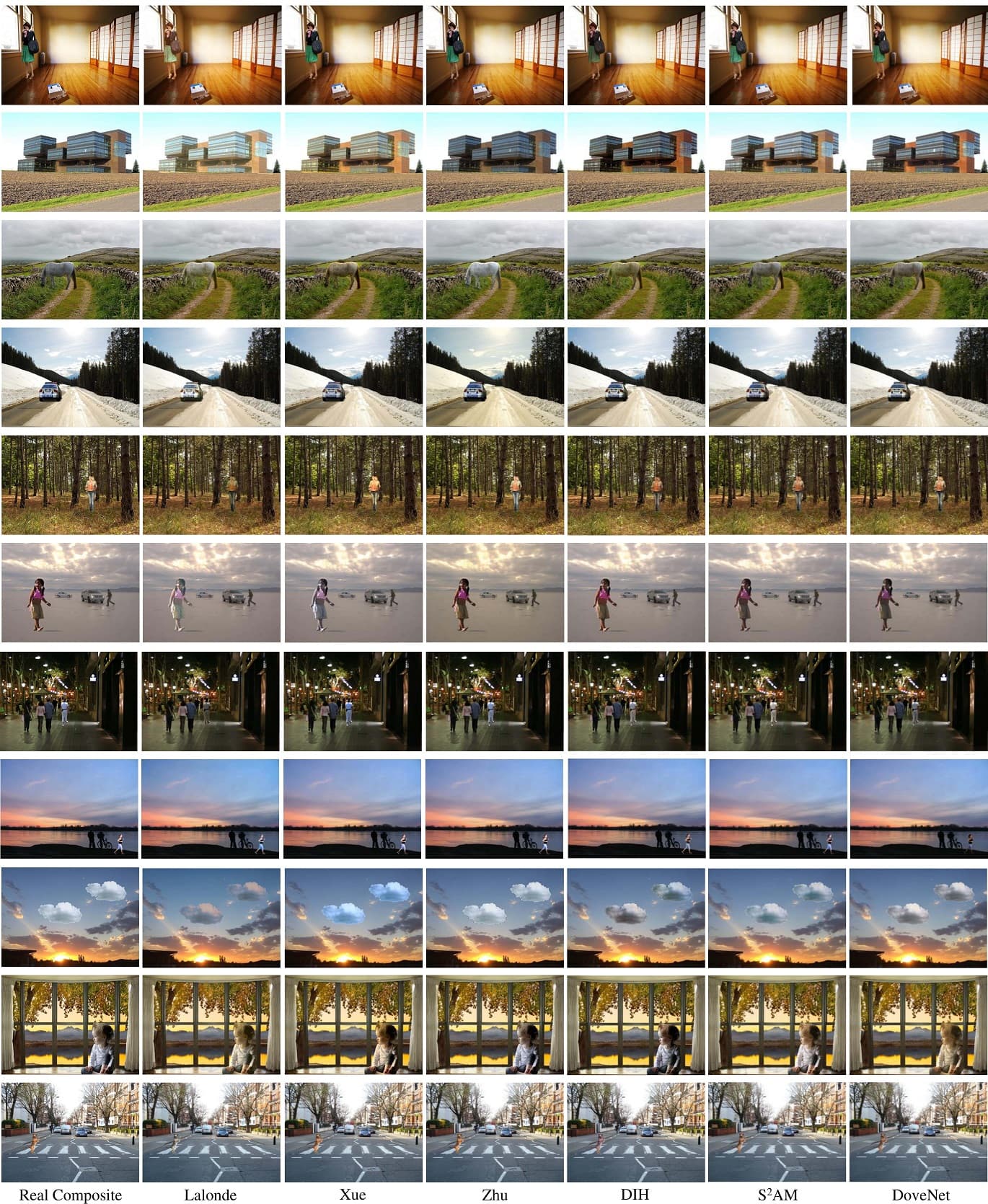}
\end{center}
   \caption{Results on real composite images, including the input composite, five state-of-the-art methods, and our proposed DoveNet. }
\label{fig:real3}
\end{figure*}

\begin{figure*}[tp!]
\begin{center}
\includegraphics[width=\linewidth]{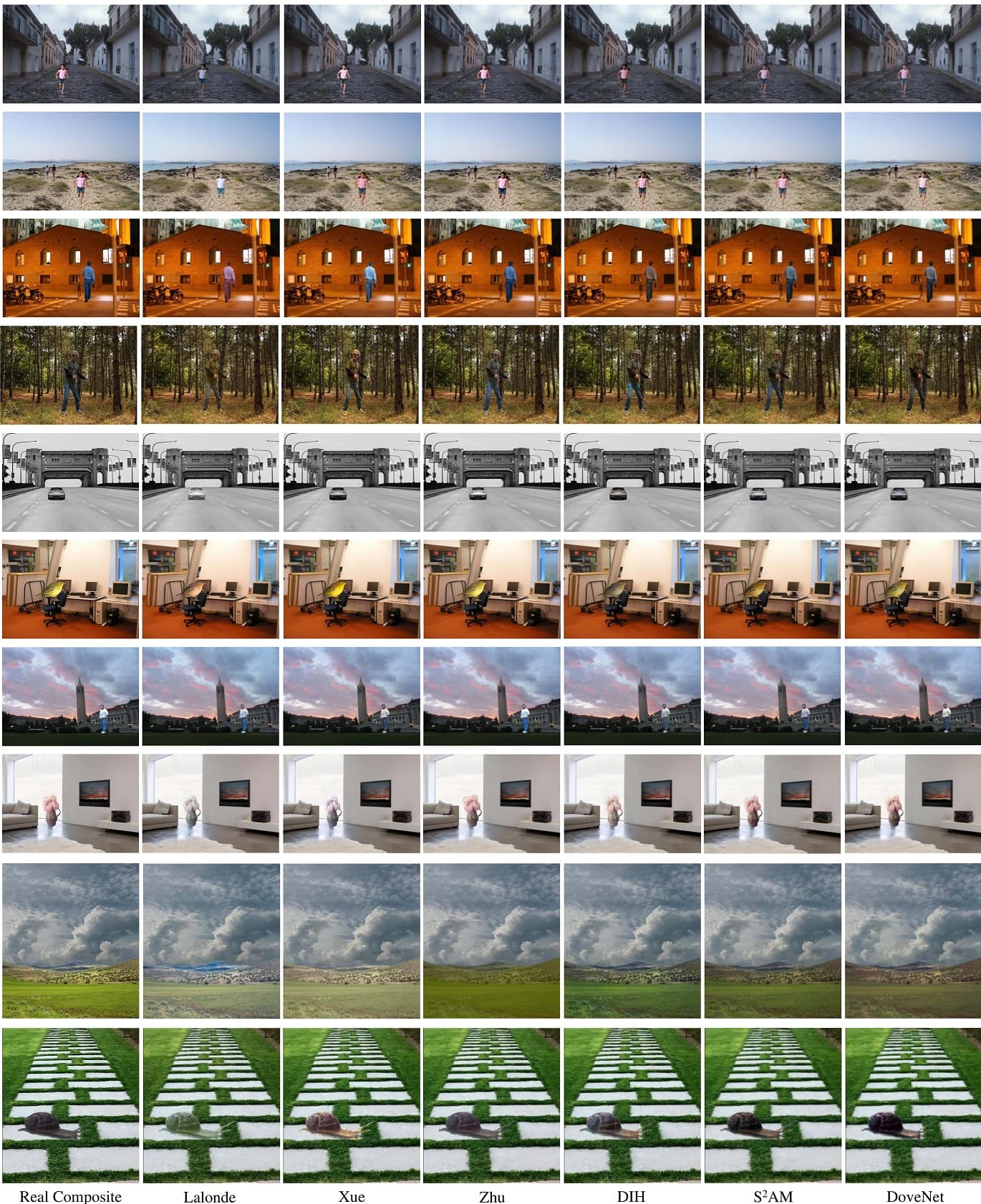}
\end{center}
   \caption{Results on real composite images, including the input composite, five state-of-the-art methods, and our proposed DoveNet. }
\label{fig:real4}
\end{figure*}

\begin{figure*}[tp!]
\begin{center}
\includegraphics[width=\linewidth]{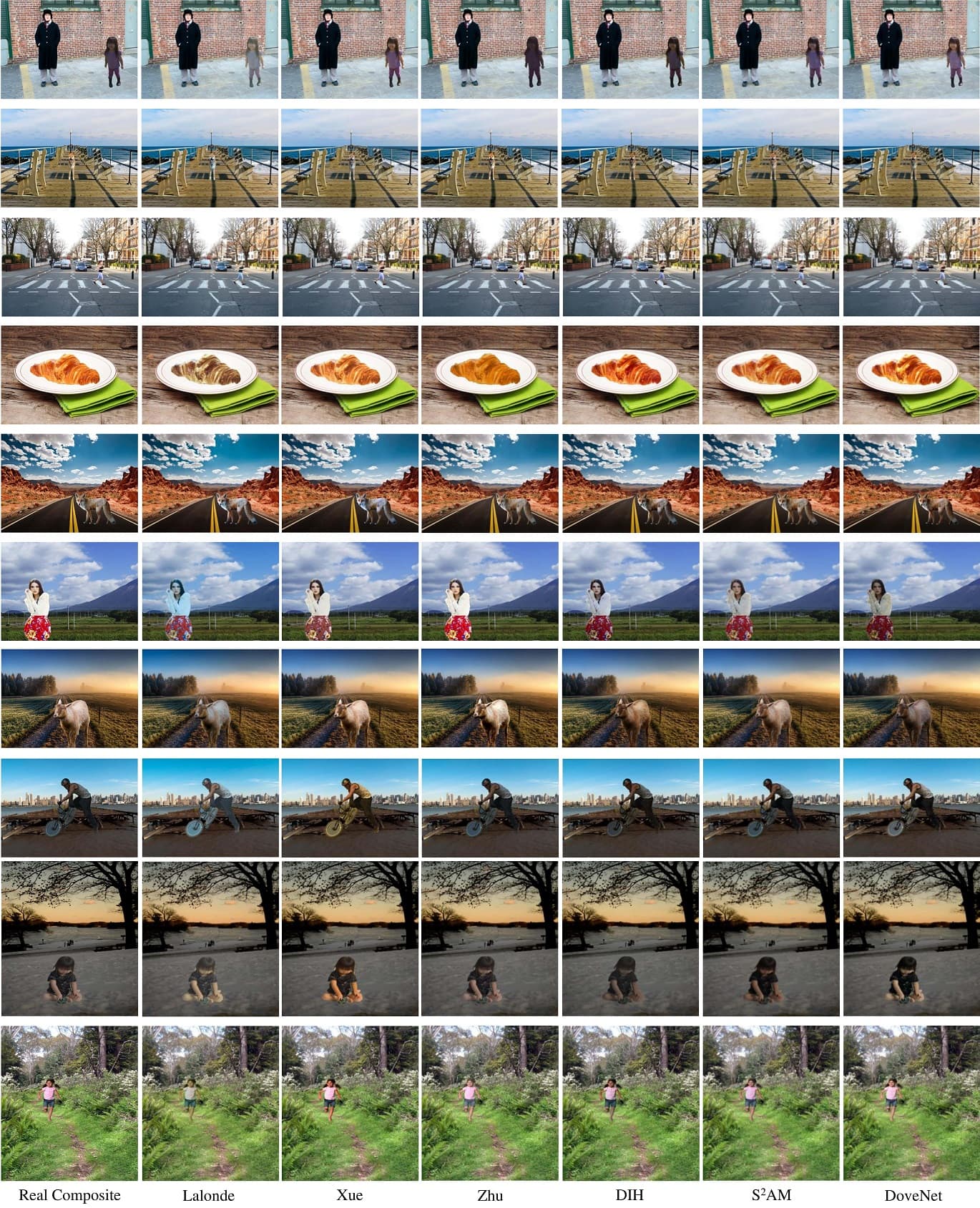}
\end{center}
   \caption{Results on real composite images, including the input composite, five state-of-the-art methods, and our proposed DoveNet. }
\label{fig:real5}
\end{figure*}

\begin{figure*}[tp!]
\begin{center}
\includegraphics[width=\linewidth]{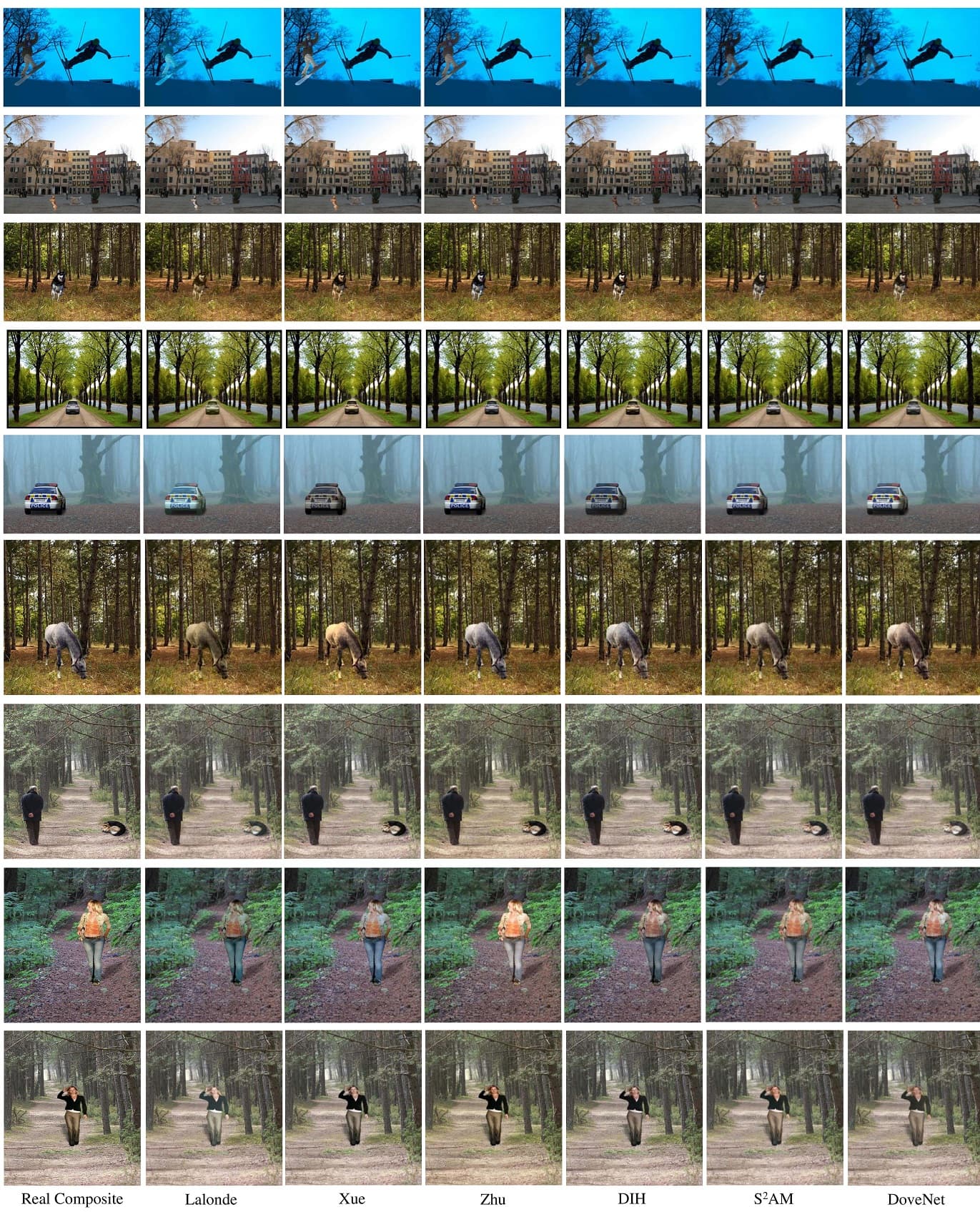}
\end{center}
   \caption{Results on real composite images, including the input composite, five state-of-the-art methods, and our proposed DoveNet. }
\label{fig:real6}
\end{figure*}

\begin{figure*}[tp!]
\begin{center}
\includegraphics[width=\linewidth]{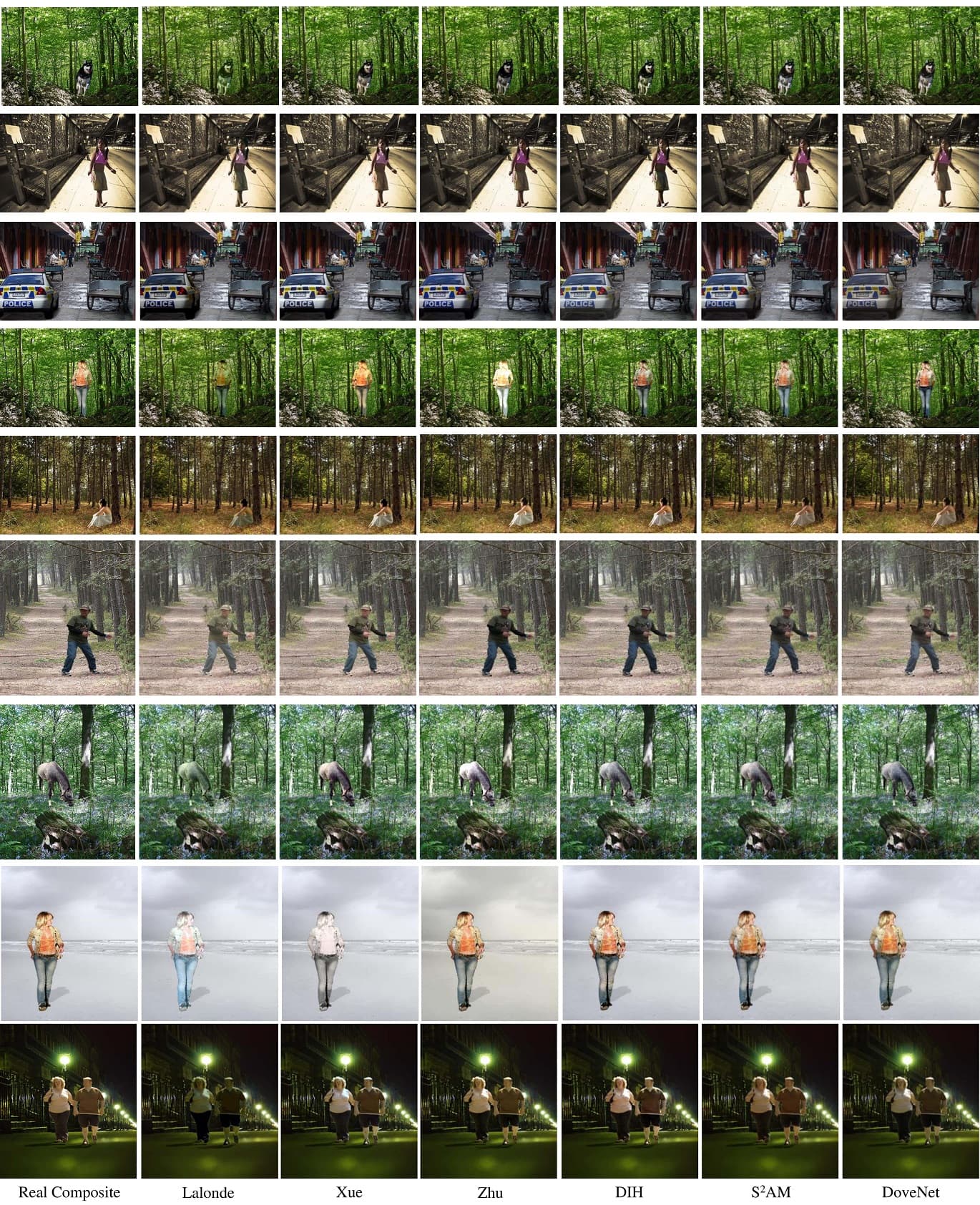}
\end{center}
   \caption{Results on real composite images, including the input composite, five state-of-the-art methods, and our proposed DoveNet. }
\label{fig:real7}
\end{figure*}

\begin{figure*}[tp!]
\begin{center}
\includegraphics[width=\linewidth]{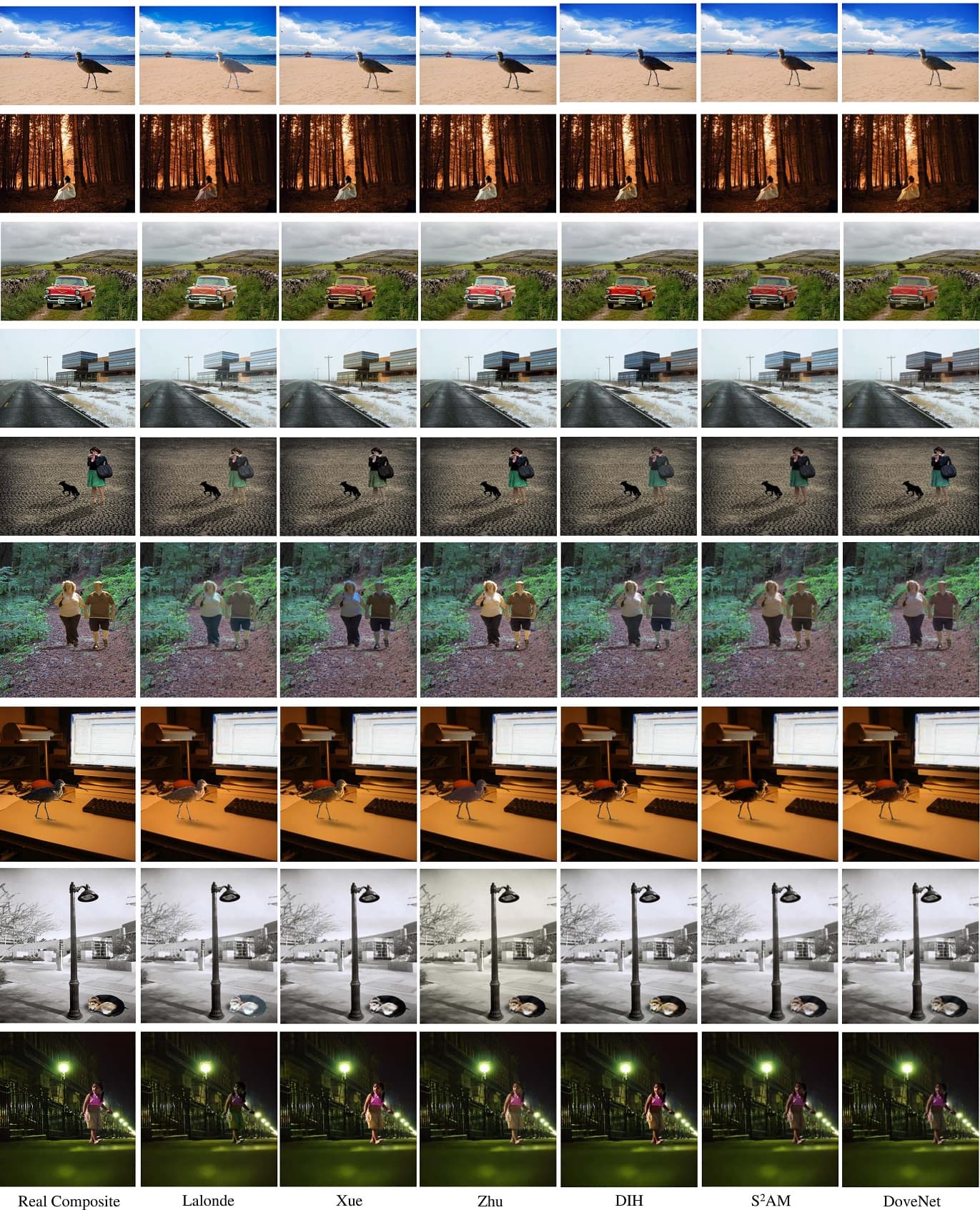}
\end{center}
   \caption{Results on real composite images, including the input composite, five state-of-the-art methods, and our proposed DoveNet. }
\label{fig:real8}
\end{figure*}

\begin{figure*}[tp!]
\begin{center}
\includegraphics[width=\linewidth]{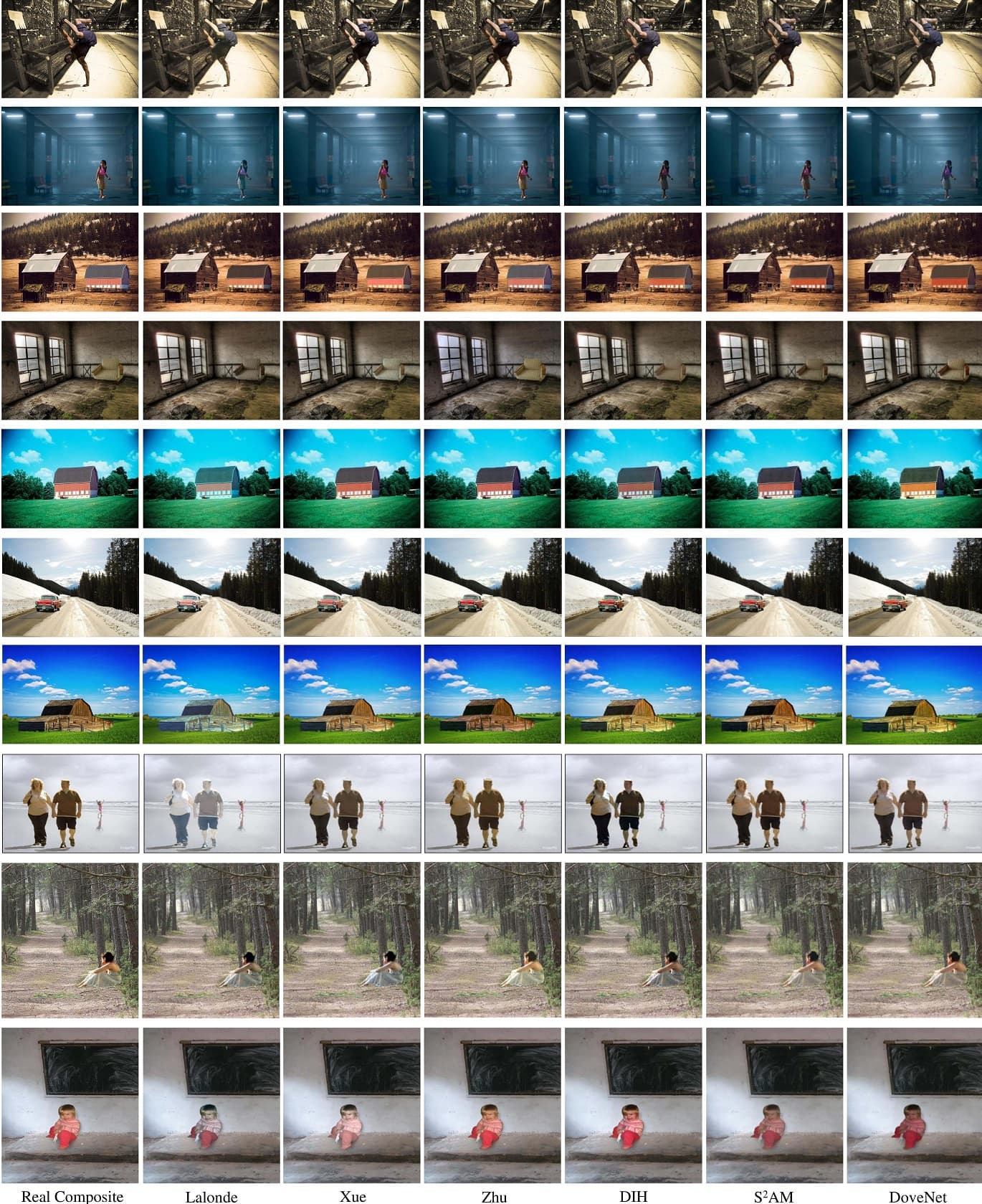}
\end{center}
   \caption{Results on real composite images, including the input composite, five state-of-the-art methods, and our proposed DoveNet. }
\label{fig:real9}
\end{figure*}

\begin{figure*}[t!]
\begin{center}
\includegraphics[width=\linewidth]{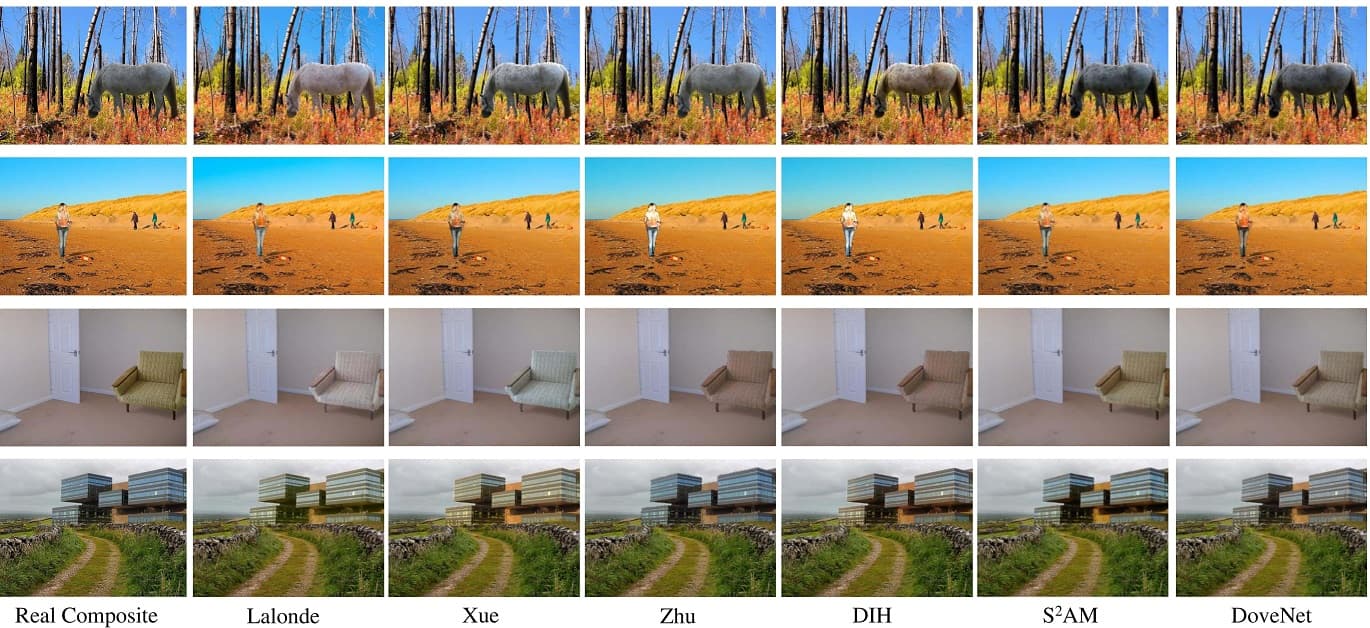}
\end{center}
  \caption{Results on real composite images, including the input composite, five state-of-the-art methods, and our proposed DoveNet. }
\label{fig:real10}
\end{figure*}

In Figure~\ref{fig:real0} to~\ref{fig:real10}, we present all results of 99 real composite images used in our user study (see Section 5.6 in the main paper), including 48 images from Xue \textit{et al.} \cite{xue2012understandingsupp} and 51 images from Tsai \textit{et al.} \cite{tsai2017deepsupp}. We compare the real composite images with harmonization results generated by our proposed method and other existing methods, including Lalonde and Efros~\cite{lalonde2007usingsupp}, Xue \textit{et al.}~\cite{xue2012understandingsupp}, Zhu \textit{et al.}~\cite{zhu2015learningsupp}, DIH ~\cite{tsai2017deepsupp}, and S$^2$AM~\cite{xiaodong2019improvingsupp}. Based on Figure~\ref{fig:real0} to \ref{fig:real10}, we can see that our proposed method could generally produce satisfactory harmonized images across various scenes and objects.

\vspace{25pt}

\pagebreak

\end{document}